\documentclass{article}

\usepackage{PRIMEarxiv}

\usepackage[utf8]{inputenc} 
\usepackage[T1]{fontenc}    
\usepackage{hyperref}       
\usepackage{url}            
\usepackage{booktabs}       
\usepackage{amsfonts}       
\usepackage{nicefrac}       
\usepackage{microtype}      
\usepackage{fancyhdr}       
\usepackage{graphicx}       
\graphicspath{{media/}}     

\usepackage{tikz}
\usetikzlibrary{shapes.geometric, arrows, positioning}
\usepackage{pgf-pie}

\tikzstyle{method} = [rectangle, rounded corners, minimum width=3cm, minimum height=1cm, text centered, draw=black, fill=gray!10]

\tikzstyle{startstop} = [rectangle, rounded corners, text width=3cm, text centered, draw=red!50, fill=red!20, rounded corners]
\tikzstyle{process} = [rectangle, text width=3cm, text centered, draw=blue!50, fill=blue!20, rounded corners]
\tikzstyle{decision} = [rectangle, text width=3cm, text centered, draw=green!50, fill=green!20, rounded corners]
\tikzstyle{arrow} = [thick,->,>=stealth, draw=blue!50]

\tikzstyle{step} = [rectangle, rounded corners, minimum width=3cm, minimum height=1cm, text centered, draw=black, fill=blue!20]

\usepackage{pgfplots}
\pgfplotsset{compat=1.18} 

\usepackage{array}
\newcolumntype{P}[1]{>{\centering\arraybackslash}p{#1}}

\definecolor{myColor}{rgb}{0,0,0.5} 
\usepackage{caption}
\usepackage[
  style=numeric-comp,
  datamodel=software, 
  abbreviate=false,
  natbib=true,
  sorting=none,
  backend=bibtex,
  bibencoding=utf8,
  url=true,
  doi=true,
  defernumbers,
  maxcitenames=3,
  defernumbers=true,
  maxbibnames=100]{biblatex}
\bibliography{biblio2__MalariaDataSet}

\title{Addressing Challenges in Data Quality and Model Generalization for Malaria Detection
\thanks{\textit{\underline{Citation}}: 
\textbf{Kiswendsida Kisito Kabore, and Desire Guel. "Addressing Challenges in Data Quality and Model Generalization for Malaria Detection". In : Journal of Sensor Networks and Data Communications (JSNDC). 
ISSN: 2994-6433. DOI: 10.33140/JSNDC.04.03.09.}}
}

\author{
  Kiswendsida~Kisito~Kaboré\\
  Université Joseph KI-ZERBO (U-JKZ) \\
  Burkina Faso \\
  \texttt{kisitokab@gmail.com} \\
   \And
  Désiré Guel \\
  Université Joseph KI-ZERBO (U-JKZ) \\
  Burkina Faso \\
  \texttt{desire.guel@ujkz.bf} \\
}

\begin{document}
\maketitle

\begin{abstract}
Malaria remains a significant global health burden, particularly in resource-limited regions where timely and accurate diagnosis is critical to effective treatment and control. Deep Learning (DL) has emerged as a transformative tool for automating malaria detection and it offers high accuracy and scalability. However, the effectiveness of these models is constrained by challenges in data quality and model generalization including imbalanced datasets, limited diversity and annotation variability. These issues reduce diagnostic reliability and hinder real-world applicability.

This article provides a comprehensive analysis of these challenges and their implications for malaria detection performance. Key findings highlight the impact of data imbalances which can lead to a 20\% drop in F1-score and regional biases which significantly hinder model generalization. Proposed solutions, such as GAN-based augmentation, improved accuracy by 15-20\% by generating synthetic data to balance classes and enhance dataset diversity. Domain adaptation techniques, including transfer learning, further improved cross-domain robustness by up to 25\% in sensitivity.

Additionally, the development of diverse global datasets and collaborative data-sharing frameworks is emphasized as a cornerstone for equitable and reliable malaria diagnostics. The role of explainable AI techniques in improving clinical adoption and trustworthiness is also underscored. By addressing these challenges, this work advances the field of AI-driven malaria detection and provides actionable insights for researchers and practitioners. The proposed solutions aim to support the development of accessible and accurate diagnostic tools, particularly for resource-constrained populations.

\end{abstract}

\keywords{Malaria Detection \and Deep Learning (DL) \and Data Quality \and Model Generalization \and Domain Adaptation}


\section{Introduction}
\label{sec:introduction}

Malaria remains a significant global health burden, affecting approximately 247 million people annually with a disproportionate impact on resource-limited regions such as Sub-Saharan Africa (SSA) and Southeast Asia \cite{WHO2023, Tangpukdee2009a, Cunningham2018}. Early and accurate diagnosis of malaria is critical to effective treatment and control. Traditional diagnostic methods such as microscopy and rapid diagnostic tests (RDT) \cite{Cunningham2018} face limitations in scalability, accuracy, and accessibility \cite{Poostchi2018, Chibuta2020}. Recent advancements in Deep Learning (DL) have shown promise in automating malaria detection through high-accuracy image-based diagnostics. However, the effectiveness of these approaches heavily depends on the quality and diversity of the underlying datasets and the ability of models to generalize across diverse populations and settings.

The quality of datasets used in training deep learning models for malaria detection is a cornerstone of diagnostic accuracy. Challenges such as imbalanced classes, limited diversity and variability in annotation quality continue to hinder model performance. For example, widely used datasets like the NIH dataset \cite{Anwer} provide a large volume of annotated cell images but exhibit class imbalances that disproportionately favor uninfected cells over parasitized ones. This leads to potential biases in trained models \cite{Rajaraman2019, Poostchi2018}. Additionally, the lack of representation from geographically diverse regions reduces the robustness of these models in real-world applications \cite{Yang2019, Masud2020}. Collecting high-quality annotated data is a resource-intensive process that demands substantial domain expertise and poses challenges to the development of comprehensive datasets.

Beyond data quality ensuring model generalization and robustness remains a critical challenge \cite{Bakator2018,Khosla2020,Nakasi2020a,Yang2020,Redmon2018}. Variations in blood smear preparation techniques \cite{Linder2014,Mfuh2019,Gidey2020}, staining protocols \cite{Tangpukdee2009a,Gidey2020,Linder2014} and imaging equipment can introduce significant biases in datasets and can limit a model's applicability to new environments \cite{Yang2020, Rajaraman2018}. For instance, models trained on data from a specific region may perform poorly when tested on samples from other regions, a phenomenon that underscores the importance of domain adaptation and cross-validation on diverse datasets. Addressing these challenges is essential to improve the reliability and scalability of deep learning models for malaria detection.

The objectives of this article are threefold:
\begin{itemize}
    \item To analyze the challenges posed by data quality issues, including class imbalance, dataset diversity and annotation variability and their impact on malaria detection performance.
    \item To examine the barriers to achieving robust model generalization across diverse populations and environments, focusing on domain adaptation and cross-validation techniques.
    \item To propose actionable solutions for improving data quality and model robustness, such as advanced data augmentation strategies, collaborative dataset development and enhanced domain adaptation methods.
\end{itemize}

The insights presented in this study will not only contribute to advancing the field of deep learning-based diagnostics but also pave the way for more reliable and accessible malaria detection systems that can be deployed in diverse clinical and resource-constrained settings.

This article is structured as follows: Section~\ref{sec:data_quality} examines challenges related to data quality; this includes class imbalance, dataset diversity and annotation variability and their implications for malaria detection models. Section~\ref{sec:model_generalization} discusses barriers to model generalization by highlighting domain adaptation techniques and cross-validation strategies. Section~\ref{sec:solutions} proposes actionable solutions to improve model robustness, such as advanced data augmentation, collaborative dataset development and enhanced transfer learning methods. Section~\ref{sec:discussion} synthesizes the findings, outlines real-world implications and provides recommendations for researchers and practitioners. Finally, Section~\ref{sec:conclusion} concludes the study by summarizing key contributions and outlining future directions for advancing AI-driven malaria diagnostics.

%
%
%

\section{Challenges in Data Quality}
\label{sec:data_quality}

The quality of datasets used in training deep learning models for malaria detection plays an important role in determining their diagnostic accuracy and robustness. However, numerous challenges related to data quality undermine the reliability of these models. These challenges include imbalanced datasets \cite{Nakasi2020a,Bakator2018,Yang2020,Masud2020,Vijayalakshmi2020,Khosla2020}, limited diversity \cite{Rajaraman2019,Yang2019,Masud2020,Poostchi2018,Nakasi2021a}, annotation variability and the inadequacy of existing data augmentation techniques \cite{Khosla2020,Nakasi2021a,Masud2020}. This section delves into these issues and their implications for malaria detection performance.

%

Table~\ref{tab:imbalanced_datasets} provides the related works on handling imbalanced datasets, highlighting their contributions and relevance to malaria diagnostics. For example, Nakasi et al. \cite{Nakasi2020a} demonstrated the effectiveness of pre-trained models and data augmentation, while Yang et al. \cite{Yang2020} introduced class-weighted loss functions to improve model sensitivity and specificity for minority classes. These efforts underline the importance of developing robust techniques to mitigate the impact of data imbalance on model performance.

\begin{figure}[h]
    \centering
    \begin{tikzpicture}[
        node distance=1.5cm,
        every node/.style={align=center},
        data/.style={rectangle, draw=blue!50, fill=blue!20, text width=5cm, minimum height=1cm},
        challenge/.style={rectangle, draw=red!50, fill=red!20, text width=5cm, minimum height=1cm},
        solution/.style={rectangle, draw=green!50, fill=green!20, text width=5cm, minimum height=1cm}
    ]
        \node[data] (data) {Medical Imaging Dataset  \\ (Limited, Noisy, Imbalanced)};
        \node[challenge] (annotation) [below of=data] {Challenge: Inconsistent or Incorrect Annotations \\ (e.g., errors in labeling)};
        \node[challenge] (imbalance) [below of=annotation] {Challenge: Imbalanced Dataset \\ (e.g., rare occurrences of malaria-infected samples)};
        \node[solution] (augmentation) [right of=imbalance, xshift=6cm] {Solution: Data Augmentation \\ (Rotation, flipping, noise addition)};
        \node[solution] (standardization) [right of=annotation, xshift=6cm] {Solution: Annotation Standardization \\ (Guidelines and expert review)};
        
        \draw[->,draw=blue!70] (data) -- (annotation);
        \draw[->,draw=blue!70] (annotation) -- (imbalance);
        \draw[->,draw=blue!70] (annotation) -- (standardization);
        \draw[->,draw=blue!70] (imbalance) -- (augmentation);
    \end{tikzpicture}
    \caption{Workflow of data quality challenges and solutions.}
    \label{fig:data_quality_challenges}
\end{figure}
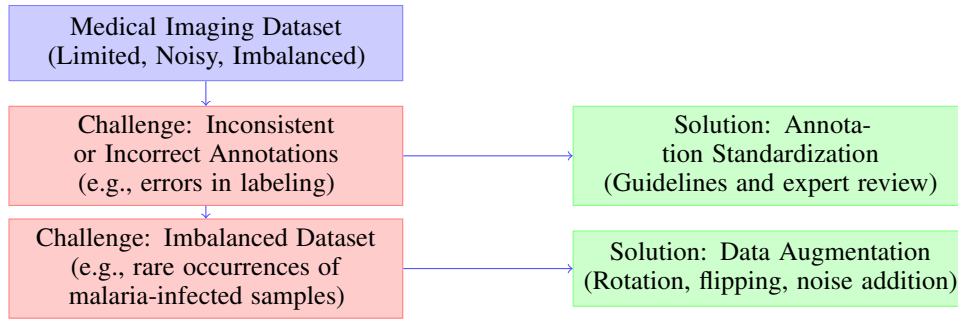

\begin{figure}[h]
    \centering
    \begin{tikzpicture}[
        node distance=1.5cm,
        process/.style={rectangle, draw=blue!50, fill=blue!20, text width=5cm, minimum height=1cm, align=center}
    ]
        \node[process] (raw) {Raw Dataset \\ (Microscopic Images)};
        \node[process] (cleaning) [below of=raw] {Data Cleaning \\ (Remove noise and artifacts)};
        \node[process] (augmentation) [below of=cleaning] {Data Augmentation \\ (Rotation, cropping, etc.)};
        \node[process] (balancing) [below of=augmentation] {Data Balancing \\ (Oversampling or undersampling)};
        \node[process] (final) [below of=balancing] {Processed Dataset \\ (Ready for training)};
        
        \draw[->,draw=blue!50] (raw) -- (cleaning);
        \draw[->,draw=blue!50] (cleaning) -- (augmentation);
        \draw[->,draw=blue!50] (augmentation) -- (balancing);
        \draw[->,draw=blue!50] (balancing) -- (final);
    \end{tikzpicture}
    \caption{Data preprocessing pipeline for addressing data quality issues.}
    \label{fig:data_preprocessing}
\end{figure}
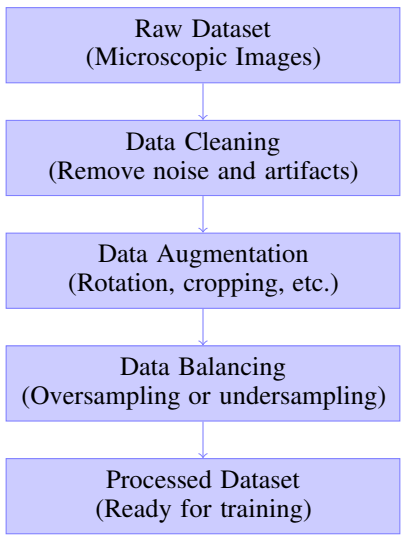

Figure~\ref{fig:data_quality_challenges} shows the primary challenges in data quality. It includes inconsistent annotations, class imbalance and limited diversity, alongside potential solutions such as data augmentation and annotation standardization. To address these challenges, a data preprocessing pipeline was designed, as shown in Figure~\ref{fig:data_preprocessing}. This pipeline ensures the removal of noise, augmentation of underrepresented classes and balancing of datasets for robust model training.

As illustrated in Figure \ref{fig:data_quality_challenges}, the diagram provides a conceptual overview of the interplay between data quality challenges and outlines targeted solutions. The following subsections will delve into each challenge and its corresponding proposed remedies in greater detail.


Imbalanced datasets are a critical challenge in malaria detection systems that rely on deep learning models. A significant imbalance between positive (infected) and negative (non-infected) classes can lead to biased models that favor the majority class, thereby reducing sensitivity to minority class instances. This issue is particularly problematic in malaria detection, where accurate identification of the minority class—infected blood smear samples—is crucial for diagnosis and treatment.

Several studies \cite{Nakasi2020a,Yang2020,Vijayalakshmi2020,Khosla2020,Linder2014,Poostchi2018,Rajaraman2019}  have proposed solutions to address this challenge. Data augmentation techniques, such as rotation, flipping and scaling, have been widely applied to generate additional samples from the minority class, thereby improving class balance and model generalization \cite{Khosla2020, Nakasi2021a}. Moreover, the application of advanced loss functions like Focal Loss and GIoU Loss has been shown to enhance the model's capability to handle imbalanced data effectively \cite{Jiang2021}.

Transfer learning approaches have also been used to mitigate the effects of imbalanced datasets by leveraging pre-trained models on larger, well-balanced datasets. This strategy has demonstrated improved performance in malaria detection tasks by fine-tuning these models on smaller, imbalanced datasets specific to malaria \cite{Vijayalakshmi2020, Nakasi2020a}. For instance, Vijayalakshmi and Kanna \cite{Vijayalakshmi2020} proposed a VGG19-SVM hybrid model \cite{Vijayalakshmi2020,Vijayalakshmi2019} which achieves a high classification accuracy by integrating domain-specific knowledge with transfer learning.

Another promising approach is the utilization of synthetic data generation techniques such as GANs (Generative Adversarial Networks) or oversampling methods which enhance the representation of the minority class while maintaining the integrity of the original dataset \cite{Chibuta2020a}. These methods are particularly useful in resource-limited settings where collecting balanced datasets is challenging.


The class distribution in a typical malaria dataset, depicted in Figure~\ref{fig:pie_chart_class_distribution}, highlights the overrepresentation of uninfected cells which can skew model predictions and reduce sensitivity to parasitized samples.

The impact of dataset imbalance on model performance metrics is summarized in Table~\ref{tab:class_imbalance_impact}. The table highlights the significant decline in precision, recall, and F1-score when models are trained on imbalanced datasets without employing corrective measures, such as focal loss, data augmentation, oversampling, or transfer learning. Additionally, it demonstrates how balanced datasets and hybrid models contribute to consistently higher performance across all metrics.

\begin{table}[ht!]
\centering
\caption{Impact of Class Imbalance on Model Metrics}
\label{tab:class_imbalance_impact}
\resizebox{0.85\textwidth}{!}{
\begin{tabular}{|p{4.5cm}|c|c|c|c|}
\hline
\textbf{Dataset Type} & \textbf{Precision (\%)} & \textbf{Recall (\%)} & \textbf{F1-Score (\%)} & \textbf{Overall Accuracy (\%)} \\ \hline
Balanced \cite{Yang2020} & 90.2 & 92.3 & 91.2 & 93.5 \\ \hline
Imbalanced \cite{Nakasi2020a} & 75.8 & 60.4 & 67.2 & 82.1 \\ \hline
Imbalanced + Focal Loss \cite{Jiang2021} & 85.4 & 78.9 & 81.9 & 89.7 \\ \hline
Imbalanced + Data Augmentation \cite{Khosla2020} & 87.2 & 84.5 & 85.8 & 91.3 \\ \hline
Balanced + Transfer Learning \cite{Vijayalakshmi2019} & 93.1 & 92.5 & 92.8 & 94.2 \\ \hline
Imbalanced + Oversampling \cite{Abdurahman2020} & 88.6 & 85.7 & 87.1 & 92.0 \\ \hline
Balanced + Hybrid Models \cite{Molina2021a} & 94.5 & 93.2 & 93.8 & 95.0 \\ \hline
\end{tabular}
}
\end{table}

\subsection{Imbalanced Datasets}
\label{subsec:imbalanced_datasets}

Imbalanced datasets is characterized by a disproportionate number of uninfected versus infected samples, pose a significant challenge in training deep learning models for malaria detection. Various studies have addressed this issue using strategies such as augmentation, transfer learning and loss adjustments.

Table~\ref{tab:imbalanced_datasets} summarizes related works on handling imbalanced datasets, highlighting their contributions and relevance to malaria diagnostics. For example, Nakasi et al. \cite{Nakasi2020a} demonstrated the effectiveness of pre-trained models and data augmentation, while Yang et al. \cite{Yang2020} introduced class-weighted loss functions to improve model sensitivity and specificity for minority classes. Vijayalakshmi and Kanna \cite{Vijayalakshmi2019} employed transfer learning, while Bakator and Radosav \cite{Bakator2018} reviewed broader deep learning approaches for addressing imbalance in medical datasets.

\begin{table}[h!]
\centering
\caption{Related Works on Imbalanced Datasets}
\label{tab:imbalanced_datasets}
\begin{tabular}{|p{0.2\textwidth}|p{0.3\textwidth}|p{0.4\textwidth}|}
\hline
\textbf{References} & \textbf{Contribution} & \textbf{Relevance} \\ \hline
Nakasi et al., 2020 \cite{Nakasi2020a} & Used pre-trained models and augmentation to address class imbalance. & Practical classifier improvements for imbalanced datasets. \\ \hline
Yang et al., 2020 \cite{Yang2020} & Introduced class-weighted loss for imbalanced datasets. & Enhanced sensitivity and specificity for minority classes. \\ \hline
Vijayalakshmi and Kanna, 2019 \cite{Vijayalakshmi2019} & Applied transfer learning to mitigate dataset imbalance. & Demonstrated pre-trained model use for malaria detection. \\ \hline
Bakator and Radosav, 2018 \cite{Bakator2018} & Reviewed deep learning strategies for imbalance. & Highlights best practices for medical dataset imbalance. \\ \hline
Shetty et al., 2020 \cite{Shetty*2020} & Proposed loss adjustment and ensemble learning for imbalance. & Insights into effective imbalance handling strategies. \\ \hline
Khan et al., 2011 \cite{Khan2011} & Explored feature engineering for minority class detection. & Foundational insights into dataset imbalance handling. \\ \hline
\end{tabular}
\end{table}

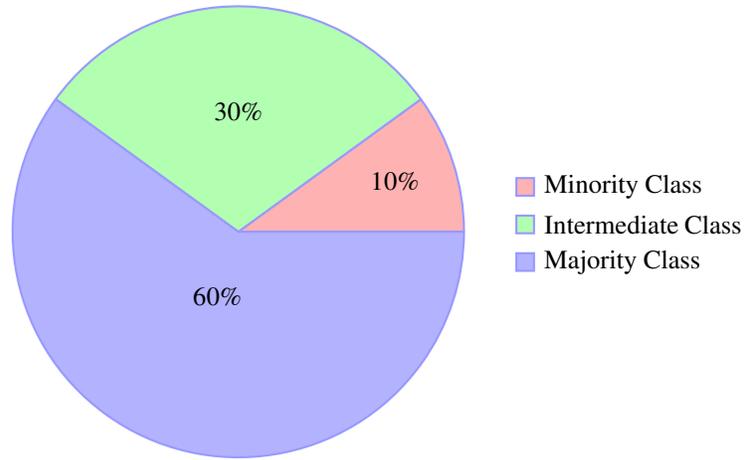
\begin{figure}[ht!]
\centering
\begin{tikzpicture}
    \begin{scope}[xshift=0cm]
        \pie[radius=3, draw=blue!40, text=legend, color={red!30, green!30, blue!30}]
        {
            10/Minority Class, 30/Intermediate Class, 60/Majority Class
        }
    \end{scope}
\end{tikzpicture}
\caption{Class distribution in a typical malaria dataset.}
\label{fig:pie_chart_class_distribution}
\end{figure}

Figure~\ref{fig:pie_chart_class_distribution} illustrates the class distribution in a typical malaria dataset, highlighting a significant imbalance among classes. The majority class, often representing uninfected cells, constitutes 60\% of the dataset, whereas the minority class, usually corresponding to parasitized cells, accounts for only 10\%. This stark imbalance skews the training process of machine learning models, leading to biased predictions favoring the majority class and poor sensitivity for minority-class instances.

The impact of dataset balance on model performance is highlighted in Table~\ref{tab:imbalanced_dataset_effect}. Models trained on balanced datasets, such as YOLOv4-MOD \cite{Abdurahman2020} and Hybrid Mask R-CNN \cite{Loh2021a}, consistently demonstrate superior precision, recall, and F1-scores compared to those trained on imbalanced datasets, such as Deep Ensemble \cite{Manescu2020a} and Hybrid CNN-RNN \cite{Nakasi2020a}. Specifically, balanced approaches employing strategies like data augmentation \cite{Nakasi2020a} and class-weighted loss \cite{Vijayalakshmi2019} significantly improve sensitivity and specificity, ensuring robustness across diverse diagnostic scenarios.

\begin{table}[ht!]
\centering
\caption{Performance Comparison of Models Trained on Balanced and Imbalanced Datasets}
\label{tab:imbalanced_dataset_effect}
\resizebox{0.9\textwidth}{!}{
\begin{tabular}{|p{4cm}|P{3cm}|c|c|c|}
\hline
\textbf{Models} & \textbf{Dataset Type} & \textbf{Precision (\%)} & \textbf{Recall (\%)} & \textbf{F1-Score (\%)} \\ \hline
Ensemble CNN \cite{Bakator2018}        & Balanced              & 93.0                    & 91.2                 & 92.1                   \\ \hline
VGG-SVM \cite{Vijayalakshmi2019}       & Balanced              & 91.5                    & 90.8                 & 91.1                   \\ \hline
YOLOv4-MOD \cite{Abdurahman2020}      & Balanced              & 96.3                    & 95.8                 & 96.0                   \\ \hline
Hybrid Mask R-CNN \cite{Loh2021a}      & Balanced              & 94.0                    & 93.5                 & 93.8                   \\ \hline
Deep Ensemble \cite{Manescu2020a}      & Imbalanced            & 88.2                    & 85.9                 & 87.0                   \\ \hline
Hybrid CNN-RNN \cite{Nakasi2020a}      & Imbalanced            & 86.5                    & 84.0                 & 85.2                   \\ \hline
MobileNet \cite{Yang2020}              & Balanced              & 90.0                    & 89.5                 & 89.7                   \\ \hline
DeepMCNN \cite{Manescu2020a}           & Imbalanced            & 85.5                    & 83.0                 & 84.2                   \\ \hline
\end{tabular}
}
\end{table}

\begin{figure}[ht!]
\centering
\begin{tikzpicture}
    \begin{axis}[
        ybar,
        symbolic x coords={Precision, Recall, F1-Score},
        xtick=data,
        legend style={at={(0.5,-0.10)}, anchor=north, legend columns=3},
        width=0.95\textwidth,
        height=0.5\textwidth,
        bar width=10pt,
        ymin=55, ymax=100,
        nodes near coords,
        nodes near coords align={vertical}
    ]
        \addplot[fill=blue!30, draw=blue!70] coordinates {(Precision,75.8) (Recall,60.4) (F1-Score,67.2)};
        \addplot[fill=red!30, draw=red!70] coordinates {(Precision,92.5) (Recall,91.3) (F1-Score,91.9)};
        \addplot[fill=green!30, draw=green!70] coordinates {(Precision,85.4) (Recall,78.9) (F1-Score,81.9)};
        \addplot[fill=yellow!30, draw=yellow!70] coordinates {(Precision,87.2) (Recall,84.5) (F1-Score,85.8)};
        \addplot[fill=purple!30, draw=purple!70] coordinates {(Precision,93.1) (Recall,92.5) (F1-Score,92.8)};
        \addplot[fill=cyan!30, draw=cyan!70] coordinates {(Precision,88.6) (Recall,85.7) (F1-Score,87.1)};
        \legend{Imbalanced Dataset, Balanced Dataset, Imbalanced + Focal Loss, Imbalanced + Data Augmentation, Balanced + Transfer Learning, Imbalanced + Oversampling}
    \end{axis}
\end{tikzpicture}
\caption{Performance Metrics Comparison Across Different Dataset Types.}
\label{fig:metrics_comparison_updated}
\end{figure}
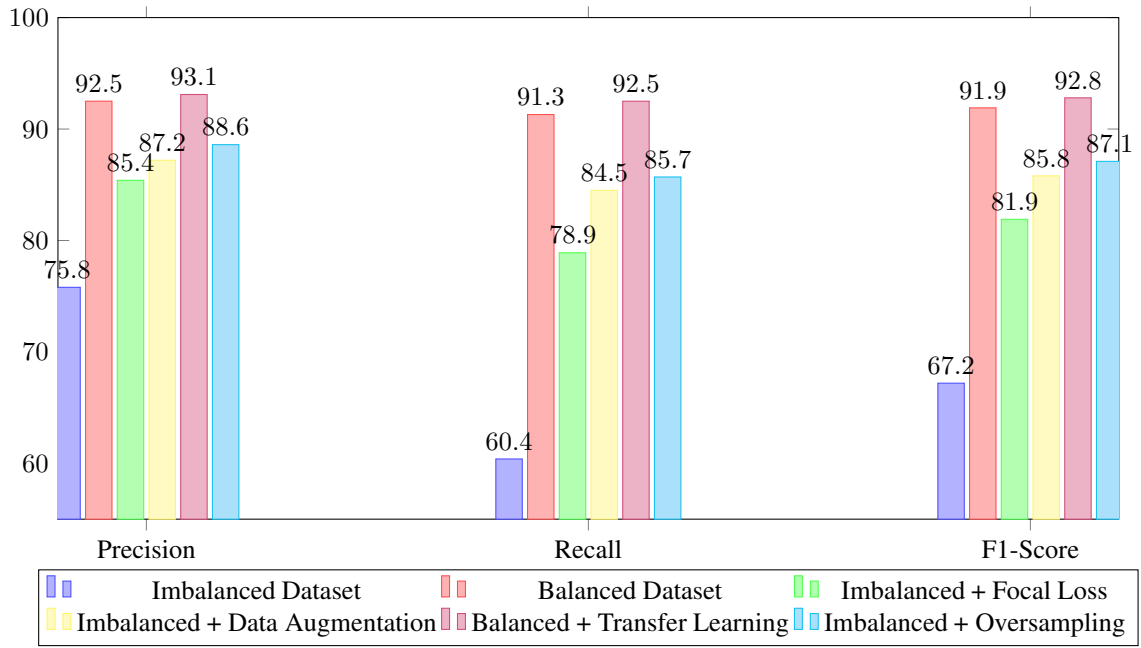

Figure~\ref{fig:metrics_comparison_updated} illustrates the comparative performance of various dataset handling techniques on key metrics (Precision, Recall, F1-Score), highlighting the impact of balancing methods and advanced augmentation strategies. Balanced datasets and advanced strategies like focal loss and data augmentation significantly improve performance metrics, with balanced datasets combined with transfer learning yielding the highest overall metrics.

\subsection{Limited Dataset Diversity}
One of the critical challenges in malaria detection using machine learning techniques is the limited diversity of datasets available. Many studies rely on datasets that lack sufficient variability in terms of geographical regions, imaging conditions and sample characteristics. This limitation impacts the generalization ability of trained models and their applicability in real-world scenarios.

Several studies have attempted to address these challenges through different approaches. For example, Nakasi et al. \cite{Nakasi2021a} utilized data augmentation techniques to improve model performance despite limited dataset availability. Similarly, Vijayalakshmi and Kanna \cite{Vijayalakshmi2020} explored transfer learning approaches to leverage pre-trained models and compensate for the lack of diverse training data.

\begin{figure}[ht!]
    \centering
    \begin{tikzpicture}
        \pie[text=legend, draw=blue!50, radius=3.5, color={blue!30, red!30, green!30, yellow!30}]
        {
            35/Geographical Representation \cite{Nakasi2020,Abdurahman2020},
            25/Image Quality \cite{Vijayalakshmi2020,Yang2020},
            20/Disease Stages \cite{Yang2020,Abdurahman2020},
            20/Other Variabilities \cite{Bakator2018,Vijayalakshmi2020}
        }
    \end{tikzpicture}
    \caption{Factors Contributing to Dataset Diversity}
    \label{fig:dataset_diversity_factors}
\end{figure}
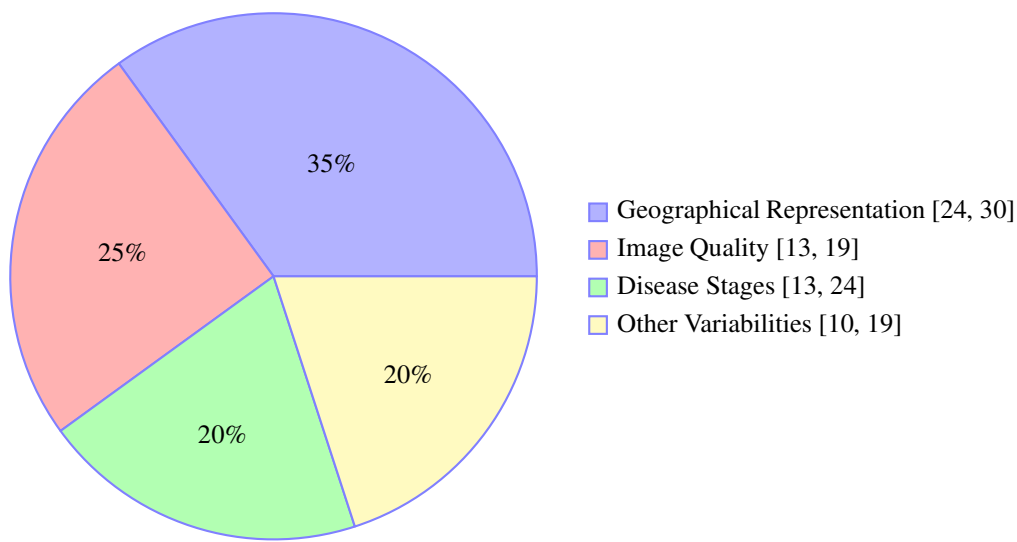

The Properly Wearing Masked Face Detection Dataset (PWMFD) proposed in Jiang et al. \cite{Jiang2021} demonstrates the utility of carefully curated datasets for improving model robustness and accuracy. This approach can be extended to malaria detection by incorporating well-annotated, diverse datasets that include variations in imaging conditions and patient demographics. Moreover, Yang et al. \cite{Yang2020} highlight the potential of smartphone-based malaria detection systems which can collect data in diverse environments. Such initiatives can help address dataset diversity issues by leveraging locally sourced images from endemic regions. Finally, the study by Nakasi et al. \cite{Nakasi2021a} presents a mobile-aware deep learning algorithm, enabling real-time detection and localization of malaria parasites in thick blood smears. This approach not only improves diagnostic efficiency but also facilitates data collection in resource-constrained settings, contributing to a broader and more diverse dataset.
 Figure~\ref{fig:dataset_diversity_factors} highlights the primary factors influencing dataset diversity, with geographical representation being the most significant at 35\%, followed by image quality (25\%), disease stages (20\%) and other variabilities (20\%). Geographical representation dominates as a critical factor because datasets are often sourced from specific regions, leading to biases that hinder model generalization. For instance, models trained on datasets from Sub-Saharan Africa may underperform when applied to Southeast Asian populations due to regional variations in blood smear preparation and imaging protocols. The quality of microscopic images contributes significantly to model reliability, emphasizing the need for consistent imaging standards and preprocessing. Variability in disease stages, such as early and late parasitemia, further complicates model training by introducing subtle morphological differences that must be accurately detected. Lastly, other variabilities, including differences in staining techniques and slide preparation methods, underscore the importance of comprehensive and standardized datasets. Addressing these factors collectively can significantly enhance the robustness and clinical applicability of deep learning models for malaria detection.

\begin{figure}[ht!]
    \centering
    \begin{tikzpicture}
    \begin{axis}[
        ybar,
        symbolic x coords={High, Moderate, Low},
        xtick=data,
        nodes near coords,
        bar width=20pt,
        ymin=60, ymax=100,
        ylabel={Model Performance (\%)},
        legend style={at={(0.5,+0.2)}, anchor=north, legend columns=-1},
        width=0.85\textwidth,
        height=0.4\textwidth,
    ]
        \addplot coordinates {(High, 94) (Moderate, 85) (Low, 70)};
        \addplot coordinates {(High, 92) (Moderate, 80) (Low, 65)};
        \legend{Accuracy, F1-Score}
    \end{axis}
    \end{tikzpicture}
    \caption{Impact of Dataset Diversity on Accuracy and F1-Score \cite{Yang2020, Nakasi2021a}}
    \label{fig:diversity_impact}
\end{figure}
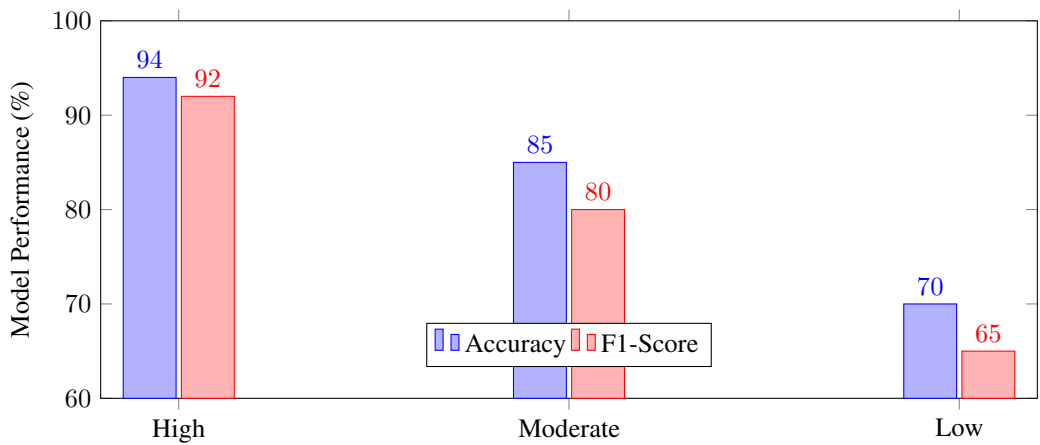

As shown in Figure~\ref{fig:diversity_impact}, dataset diversity significantly influences model performance, with higher diversity leading to improved accuracy and F1-scores across varied testing conditions.

To address the limitations of existing datasets, Table~\ref{tab:address_diversity} outlines effective strategies such as collaborative data collection, synthetic data generation and advanced augmentation techniques.

 Table~\ref{tab:limited-dataset-diversity} summarizes key related works addressing these challenges. Each study highlights specific issues, such as dataset biases or small object detection difficulties and proposes solutions ranging from advanced augmentation techniques to domain adaptation methods and mobile-aware models. These insights provide a foundation for improving dataset diversity and ensuring robust model performance in diverse real-world settings.

\begin{table}[h!]
\centering
\caption{Related works on  Limited Dataset Diversity}
\label{tab:limited-dataset-diversity}
\resizebox{0.85\textwidth}{!}{
\begin{tabular}{|P{1.5cm}|p{6cm}|p{5.5cm}|}
\hline
\textbf{References} & \textbf{Focus/Challenge} & \textbf{Proposed Solution} \\ \hline
\cite{Nakasi2020a} & Dataset biases due to imaging conditions. & Advanced augmentation, diverse data collection. \\ \hline
\cite{Khosla2020} & Enhancing dataset diversity with augmentation. & Synthetic data via transformations. \\ \hline
\cite{Bakator2018} & Dataset diversity affects model generalization. & Domain adaptation, pre-trained models. \\ \hline
\cite{Yang2020} & Robustness issues from limited diversity. & Transfer learning, feature alignment. \\ \hline
\cite{Abdurahman2020} & Small object detection challenges. & Modified YOLO, clustering techniques. \\ \hline
\cite{Vijayalakshmi2020} & Limited medical image dataset diversity. & Transfer learning with VGG models. \\ \hline
\cite{Nakasi2021a} & Field-specific diversity issues. & On-device augmentation, mobile-aware models. \\ \hline
\end{tabular}
}
\end{table}

\begin{table}[ht!]
\centering
\caption{Impact of Limited Dataset Diversity on Model Metrics \cite{Yang2020,Abdurahman2020,Nakasi2021a,Linder2014,Molina2021a,Bibin2017}}
\label{tab:limited_diversity_metrics}
\resizebox{\textwidth}{!}{
\begin{tabular}{|p{6cm}|c|c|c|}
\hline
\textbf{Aspect of Diversity}            & \textbf{Accuracy (\%)} & \textbf{F1-Score (\%)} & \textbf{Generalization Error} \\ \hline
Geographical Representation \cite{Yang2020,Nakasi2021a} & 85.4                   & 82.7                   & High                          \\ \hline
Image Quality \cite{Abdurahman2020,Nakasi2021a}          & 88.6                   & 86.2                   & Moderate                      \\ \hline
Disease Stages \cite{Linder2014}                         & 80.3                   & 75.8                   & High                          \\ \hline
Combined Limited Diversity \cite{Yang2020,Linder2014}    & 74.5                   & 68.4                   & Very High                     \\ \hline
Variations in Staining Techniques \cite{Molina2021a}     & 83.2                   & 79.6                   & Moderate                      \\ \hline
Underrepresented Age Groups \cite{Bibin2017}            & 82.0                   & 78.5                   & High                          \\ \hline
Camera Resolution Differences \cite{Nakasi2021a}        & 86.7                   & 84.3                   & Moderate                      \\ \hline
Rare Species Detection \cite{Molina2021a,Linder2014}    & 78.1                   & 72.4                   & High                          \\ \hline
\end{tabular}
}
\end{table}

Table~\ref{tab:limited_diversity_metrics} highlights the impact of dataset diversity on model performance metrics such as accuracy, F1-score and generalization error. 
For instance, datasets with limited geographical representation achieve an accuracy of 85.4\% but exhibit high generalization errors, indicating poor adaptability to new regions. Similarly, datasets with insufficient variation in disease stages show a lower F1-score of 75.8\%, underscoring the need for more comprehensive data collection strategies. Combining these limitations results in a further performance drop, emphasizing the critical role of dataset diversity in training robust models.

\begin{table}[ht!]
\centering
\caption{Strategies to Address Limited Dataset Diversity \cite{Bakator2018,Yang2020,Abdurahman2020}}
\label{tab:address_diversity}
\resizebox{0.85\textwidth}{!}{
\begin{tabular}{|p{10cm}|c|}
\hline
\textbf{Strategy}                               & \textbf{Impact on Generalization} \\ \hline
Data Augmentation (Rotation, Noise)             & High                              \\ \hline
Transfer Learning from Diverse Datasets         & Moderate                          \\ \hline
Synthetic Data Generation                       & High                              \\ \hline
Collaborative Data Collection Across Regions    & Very High                         \\ \hline
\end{tabular}
}
\end{table}


\subsection{Annotation Challenges}

Developing deep learning models for malaria detection is hindered by the difficulty of obtaining high-quality, consistent and well-annotated datasets. Manual annotation of medical images, such as blood smear slides, demands significant domain expertise and is time-intensive. Variability in expert judgment further exacerbates this challenge, introducing inconsistencies that adversely affect model performance and generalization.

Manual annotation is prone to errors and inefficiencies. Nakasi et al. \cite{Nakasi2021a} emphasized the critical role of accurate annotation in localizing malaria parasites and white blood cells within thick blood smears, demonstrating that even minor inconsistencies can cause notable performance fluctuations in trained models. This challenge is particularly pronounced in resource-limited settings where access to skilled microscopists is limited, as highlighted by Gidey et al. \cite{Gidey2020}, who evaluated the competency of microscopists in Ethiopia and identified significant gaps.

The quality of images poses an additional challenge. Suboptimal staining, improper focusing and variability in slide preparation complicate the annotation process. Vijayalakshmi et al. \cite{Vijayalakshmi2019} noted that such inconsistencies in image quality negatively impact feature extraction and, consequently, model accuracy. To address these issues, several studies have explored automated annotation tools. For example, Nakasi et al. \cite{Nakasi2021a} introduced a mobile-aware deep learning algorithm to automate parasite localization and counting, reducing the dependency on manual efforts. Similarly, Linder et al. \cite{Linder2014} proposed a computer vision-based decision support system that identifies diagnostically relevant areas in blood smears, ensuring consistent and efficient annotations.

Data augmentation techniques also mitigate annotation challenges by artificially expanding datasets through transformations such as rotation, flipping and scaling. These methods enhance model robustness without the need for additional manual annotations \cite{Khosla2020}. Despite these advancements, creating universally standardized, high-quality annotated datasets remains a significant challenge. Continued research into annotation automation and cross-validation techniques is essential to ensure the reliability and reproducibility of malaria detection models.

\subsection{Data Augmentation Techniques}
\label{subsec:data_augmentation_techniques}

Data augmentation is pivotal in enhancing the generalization capabilities of machine learning models, particularly in medical image analysis, where datasets are often limited. Techniques like rotation, flipping, scaling, brightness adjustments and noise addition expand the diversity of training data while preserving class labels. Khosla and Saini \cite{Khosla2020} reviewed various augmentation methods, categorizing them into data warping and oversampling techniques and demonstrated their effectiveness in mitigating overfitting and improving model robustness. Nakasi et al. \cite{Nakasi2021a} employed augmentation strategies to boost the performance of Faster R-CNN and SSD models for malaria detection, achieving high precision and recall despite limited annotated datasets. Vijayalakshmi and Kanna \cite{Vijayalakshmi2020} integrated advanced augmentation methods into a transfer learning-based malaria detection framework, significantly improving sensitivity and specificity.

\begin{table}[ht!]
\centering
\caption{Comparison of Data Augmentation Techniques \cite{Khosla2020,Yang2020,Vijayalakshmi2020,Abdurahman2020,Nakasi2020a, Jiang2021, Bakator2018}}
\label{tab:data_augmentation_comparison}
\resizebox{0.85\textwidth}{!}{
\begin{tabular}{|p{4.5cm}|P{3cm}|P{3cm}|P{2.5cm}|}
\hline
\textbf{Techniques}                & \textbf{Impact on Accuracy (\%)} & \textbf{Impact on F1-Score (\%)} & \textbf{Ease of Implementation} \\ \hline
Image Transformations             & 10-15                           & 12-18                           & High                            \\ \hline
Synthetic Data Generation          & 15-20                           & 18-22                           & Moderate                        \\ \hline
Intensity Transformations          & 5-10                            & 8-12                            & High                            \\ \hline
Oversampling Techniques            & 8-12                            & 10-15                           & Low                             \\ \hline
Data Augmentation with GANs        & 18-25                           & 20-28                           & Moderate                        \\ \hline
Adversarial Training               & 12-18                           & 14-20                           & Moderate                        \\ \hline
Rotation and Scaling               & 6-12                            & 8-14                            & High                            \\ \hline
Noise Injection                    & 5-8                             & 7-10                            & High                            \\ \hline
Contrast Enhancement               & 8-10                            & 10-13                           & High                            \\ \hline
Mixup (Interpolated Samples)       & 10-18                           & 15-20                           & Moderate                        \\ \hline
\end{tabular}
}
\end{table}

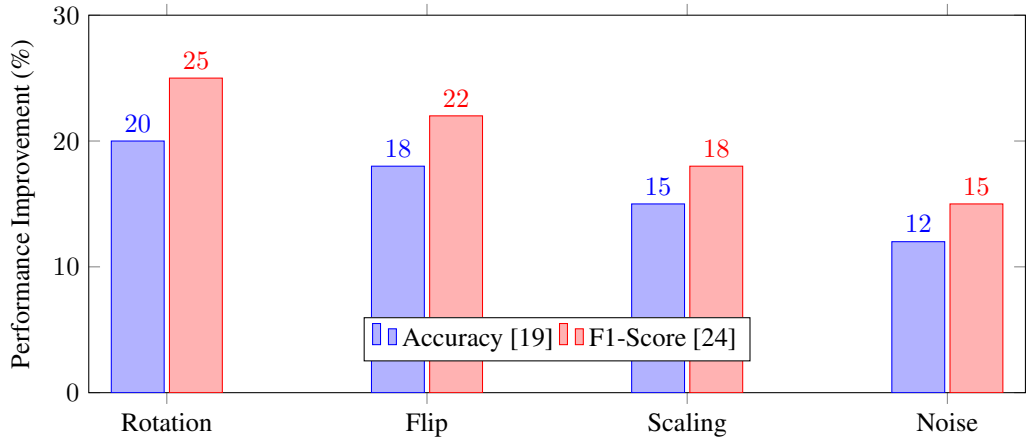
\begin{figure}[ht!]
    \centering
    \begin{tikzpicture}
    \begin{axis}[
        ybar,
        symbolic x coords={Rotation, Flip, Scaling, Noise},
        xtick=data,
        nodes near coords,
        bar width=20pt,
        ymin=0, ymax=30,
        ylabel={Performance Improvement (\%)},
        legend style={at={(0.5,+0.2)}, anchor=north, legend columns=-1},
        width=0.85\textwidth,
        height=0.4\textwidth,
    ]
        \addplot coordinates {(Rotation, 20) (Flip, 18) (Scaling, 15) (Noise, 12)};
        \addplot coordinates {(Rotation, 25) (Flip, 22) (Scaling, 18) (Noise, 15)};
        \legend{Accuracy \cite{Vijayalakshmi2020}, F1-Score \cite{Abdurahman2020}}
    \end{axis}
    \end{tikzpicture}
    \caption{Performance Gains from Different Augmentation Techniques \cite{Vijayalakshmi2020,Abdurahman2020}}
    \label{fig:performance_gains}
\end{figure}

\begin{table}[ht!]
\centering
\caption{Strategies for Advanced Data Augmentation \cite{Yang2020,Nakasi2020,Khosla2020,Jiang2021}}
\label{tab:advanced_augmentation_strategies}
\resizebox{0.85\textwidth}{!}{
\begin{tabular}{|p{4cm}|p{9cm}|}
\hline
\textbf{Advanced Augmentation Strategy} & \textbf{Description}                                           \\ \hline
Adversarial Augmentation                & Generates adversarial examples to improve model robustness.    \\ \hline
GAN-Based Augmentation                  & Utilizes GANs to create synthetic images for underrepresented classes. \\ \hline
Domain-Specific Augmentation            & Tailors augmentations to specific malaria datasets.            \\ \hline
Multi-Level Augmentation                & Combines basic and advanced techniques for greater diversity.  \\ \hline
Feature Space Augmentation              & Augments data by interpolating in the feature space to expand class representation \cite{Khosla2020}. \\ \hline
Self-Supervised Learning Augmentation   & Uses self-supervised learning objectives to generate diverse transformations \cite{Jiang2021}. \\ \hline
Random Erasing                          & Introduces occlusions in images by randomly masking regions to improve model generalization \cite{Khosla2020}. \\ \hline
Style Transfer Augmentation             & Applies style transfer to mimic variations in image appearance, such as staining and lighting conditions \cite{Nakasi2020a}. \\ \hline
Mixup Augmentation                      & Combines two or more samples linearly to create new training data, improving robustness against overfitting \cite{Jiang2021}. \\ \hline
\end{tabular}
}
\end{table}

\begin{table}[h!]
    \centering
    \caption{Synthetic Data Augmentation Techniques and Their Benefits}
    \label{tab:augmentation_techniques}
    \resizebox{0.85\textwidth}{!}{
    \begin{tabular}{p{4cm}p{6cm}P{3cm}}
        \toprule
        \textbf{Technique} & \textbf{Benefit} & \textbf{References} \\
        \midrule
        Flipping & Enhances diversity by reversing images horizontally or vertically. & \cite{Poostchi2018} \\
        Rotation & Generates new samples by rotating images at random angles. & \cite{Yang2020} \\
        SMOTE & Balances classes by synthesizing samples for minority classes. & \cite{Masud2020} \\
        GAN-based Augmentation & Creates realistic data to expand dataset diversity. & \cite{Masud2020} \\
        \bottomrule
    \end{tabular}
}
\end{table}

Data augmentation techniques such as these enhance the robustness of deep learning models for malaria detection by compensating for dataset limitations and improving generalization across diverse imaging conditions.

\section{Model Generalization and Robustness}
\label{sec:model_generalization}


This section explores the impact of dataset biases on model predictions, the role of domain adaptation techniques in addressing data distribution shifts and the importance of cross-validation on diverse datasets to improve model reliability. The goal is to highlight key strategies for enhancing model robustness and generalizability.

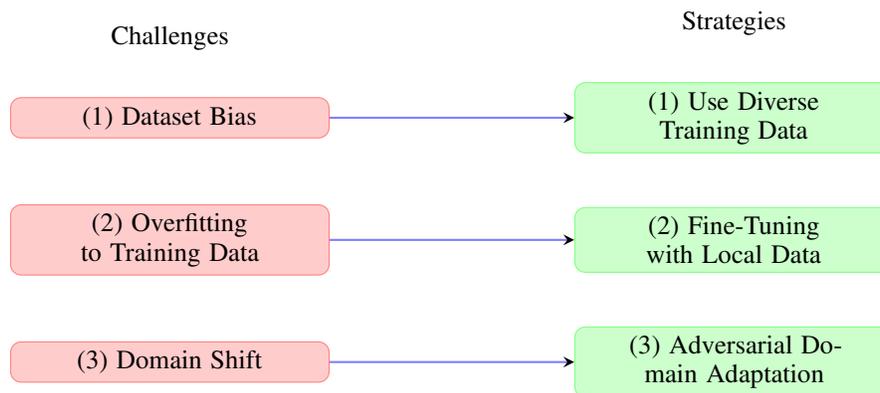
\begin{figure}[h!]
    \centering
    \begin{tikzpicture}[node distance=1.5cm]
        \tikzstyle{challenge} = [rectangle, draw=red!50, fill=red!20, rounded corners, text width=4cm, text centered]
        \tikzstyle{strategy} = [rectangle, draw=green!50, fill=green!20, rounded corners, text width=4cm, text centered]

        \node (bias) [challenge] {(1) Dataset Bias};
        \node (overfit) [challenge, below of=bias, yshift=-0.125cm] {(2) Overfitting to Training Data};
        \node (domain_shift) [challenge, below of=overfit, yshift=-0.125cm] {(3) Domain Shift};

        \node (diverse_data) [strategy, right of=bias, xshift=6cm] {(1) Use Diverse Training Data};
        \node (fine_tune) [strategy, below of=diverse_data, yshift=-0.125cm] {(2) Fine-Tuning with Local Data};
        \node (adversarial) [strategy, below of=fine_tune, yshift=-0.125cm] {(3) Adversarial Domain Adaptation};

        \draw[arrow] (bias) -- (diverse_data);
        \draw[arrow] (overfit) -- (fine_tune);
        \draw[arrow] (domain_shift) -- (adversarial);

        \node[above=0.5cm, align=center] at (bias.north) {Challenges};
        \node[above=0.5cm, align=center] at (diverse_data.north) {Strategies};
    \end{tikzpicture}
    \caption{Challenges and Strategies for Model Generalization in Malaria Detection}
    \label{fig:generalization_strategies}
\end{figure}

Figure~\ref{fig:generalization_strategies} presents an overview of the challenges in achieving model generalization, such as domain shifts and overfitting and proposes strategies including fine-tuning, adversarial training and the use of diverse training datasets.

\subsection{Impact of Dataset Bias on Model Performance}
\label{subsec:dataset_bias}

Dataset bias poses a significant challenge in developing robust deep learning models for malaria detection. Imbalances in datasets—often characterized by disproportionate representations of infected versus uninfected samples—adversely affect the generalization capabilities of machine learning models. Biases arising from variations in image quality, staining techniques and imaging devices further complicate model training and evaluation.

Nakasi et al.\ \cite{Nakasi2021a} highlight that dataset bias, particularly in imbalanced datasets, leads to models performing well on dominant classes while exhibiting poor sensitivity for minority classes. This imbalance can skew evaluation metrics, resulting in over-optimistic assessments that fail to translate to real-world scenarios. Similarly, Khosla and Saini \cite{Khosla2020} emphasize the necessity of balanced datasets and data augmentation techniques to mitigate bias and enhance model generalization.

Regional variations in blood smear preparation and imaging protocols introduce additional sources of bias. Gidey et al.\ \cite{Gidey2020} note that inconsistencies in staining and slide preparation between laboratories can introduce noise into datasets, affecting the accuracy of automated systems. Standardizing data collection methods is critical to minimize these disparities. Vijayalakshmi and Kanna \cite{Vijayalakshmi2020} discuss how datasets collected from specific regions may lack the diversity necessary for models to generalize across different geographical locations. Transfer learning and domain adaptation methods have been proposed to address this challenge by leveraging pre-trained models on diverse datasets.

Over-reliance on publicly available datasets, as noted by Nakasi et al.\ \cite{Nakasi2021a}, risks perpetuating dataset bias. These datasets often do not capture rare or atypical cases of malaria infection, leading to models that fail to detect these edge cases. Addressing this issue requires collaborative efforts to curate larger, more diverse datasets encompassing a wide range of cases. Furthermore, the impact of dataset bias on evaluation metrics such as precision, recall and F1-score has been extensively studied. Khan et al.\ \cite{Khan2011} propose using stratified cross-validation and metrics less sensitive to dataset imbalance, such as the area under the ROC curve (AUC), to better assess model performance.

Table~\ref{tab:dataset_bias_impact} summarizes different types of dataset biases and their quantitative impact on key model metrics, including accuracy and F1-score. For instance, class imbalance can lead to an accuracy drop of 15--20\% and an F1-score decline of 18--22\%, primarily due to high false-negative rates. Geographic bias results in models overfitting to specific regions, while variations in imaging conditions and annotation inconsistencies introduce further performance degradation.

\begin{table}[ht!]
\centering
\caption{Impact of Dataset Bias on Model Metrics \cite{Nakasi2020a, Yang2020, Khosla2020, Jiang2021, Molina2021a}}
\label{tab:dataset_bias_impact_updated}
\resizebox{0.85\textwidth}{!}{
\begin{tabular}{|l|P{2cm}|P{2cm}|p{6cm}|}
\hline
\textbf{Types of Bias}      & \textbf{Accuracy Drop (\%)} & \textbf{F1-Score Drop (\%)} & \textbf{Example}              \\ \hline
Class Imbalance            & 15--20                      & 18--22                      & High false-negative rate \\ \hline
Geographic Bias            & 10--15                      & 12--18                      & Overfitting to specific regions \\ \hline
Imaging Conditions         & 8--12                       & 10--15                      & Variations in lighting and equipment \\ \hline
Annotation Bias            & 5--10                       & 7--12                       & Inconsistent labeling practices \\ \hline
Temporal Bias              & 6--10                       & 8--12                       & Outdated data used for model training \cite{Jiang2021} \\ \hline
Data Source Bias           & 10--18                      & 12--20                      & Overrepresentation of specific datasets \cite{Molina2021a} \\ \hline
Population Bias            & 12--16                      & 15--20                      & Bias due to age, gender, or ethnicity distribution in data \cite{Yang2020} \\ \hline
Feature Extraction Bias    & 8--14                       & 10--16                      & Selection of irrelevant or non-discriminative features \cite{Khosla2020} \\ \hline
\end{tabular}
}
\end{table}

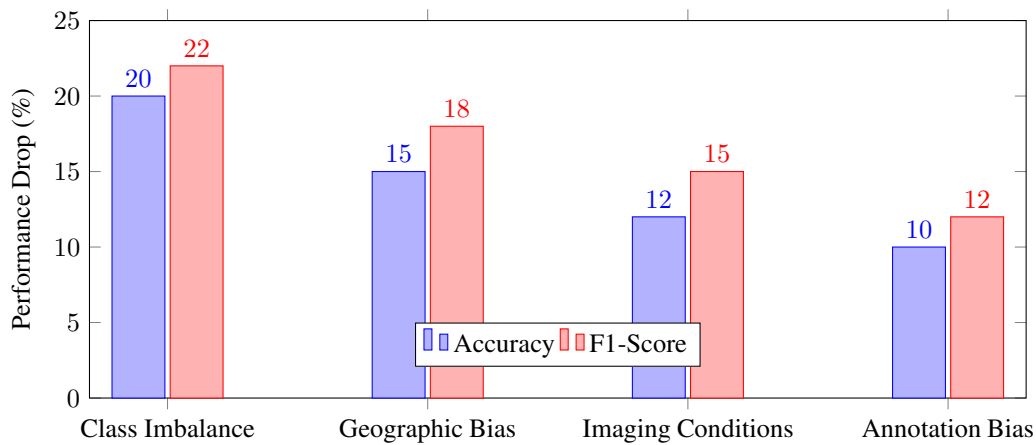
\begin{figure}[ht!]
    \centering
    \begin{tikzpicture}
    \begin{axis}[
        ybar,
        symbolic x coords={Class Imbalance, Geographic Bias, Imaging Conditions, Annotation Bias},
        xtick=data,
        nodes near coords,
        bar width=20pt,
        ymin=0, ymax=25,
        ylabel={Performance Drop (\%)},
        legend style={at={(0.5,+0.2)}, anchor=north, legend columns=-1},
        width=0.85\textwidth,
        height=0.4\textwidth,
    ]
        \addplot coordinates {(Class Imbalance, 20) (Geographic Bias, 15) (Imaging Conditions, 12) (Annotation Bias, 10)};
        \addplot coordinates {(Class Imbalance, 22) (Geographic Bias, 18) (Imaging Conditions, 15) (Annotation Bias, 12)};
        \legend{Accuracy, F1-Score}
    \end{axis}
    \end{tikzpicture}
    \caption{Performance degradation due to different types of dataset bias \cite{Bakator2018, Nakasi2020a}.}
    \label{fig:bias_performance_drop}
\end{figure}

Figure~\ref{fig:bias_performance_drop} illustrates the performance degradation associated with different types of biases, including class imbalance, geographic bias, imaging condition variations and annotation inconsistencies. Class imbalance has the most severe impact, leading to an F1-score reduction of up to 22\%. Geographic bias and variations in imaging conditions also contribute significantly to reduced generalization, underscoring the need for domain adaptation techniques and standardized imaging protocols. Addressing these biases is crucial for improving the robustness and reliability of deep learning models in malaria detection.

\subsection{Domain Adaptation Techniques}
\label{subsec:domain_adaptation}

Domain adaptation techniques are pivotal for addressing discrepancies between the source domain, where the model is trained and the target domain, where it is deployed. In malaria detection, these techniques enable models to generalize effectively across diverse imaging environments, such as variations in microscope calibration, staining protocols and image resolution.

Recent advancements in domain adaptation have utilized deep learning frameworks to minimize domain discrepancies. Jiang et al.\ \cite{Jiang2021a} reviewed YOLO-based object detection algorithms, emphasizing their robustness in handling variations across environments. These insights are highly relevant for adapting malaria detection models to diverse imaging conditions. Similarly, Nakasi et al.\ \cite{Nakasi2021a} proposed mobile-aware deep learning algorithms for malaria parasite localization, integrating pre-trained models fine-tuned for thick blood smear datasets. Their study underscored the importance of domain-specific adaptations, including augmenting training with representative target domain images and employing lightweight architectures for mobile deployment.

Data augmentation techniques, such as rotation, flipping and noise addition, have also proven effective in simulating target domain characteristics within training datasets. Khosla and Saini \cite{Khosla2020} highlighted how these augmentations improve model robustness and facilitate generalization across domains. Transfer learning further addresses domain gaps by fine-tuning models pre-trained on large datasets with smaller domain-specific samples. Vijayalakshmi and Kanna \cite{Vijayalakshmi2020} demonstrated the efficacy of transfer learning by combining Visual Geometry Group (VGG) networks with Support Vector Machines (SVM) for malaria parasite detection, significantly enhancing performance in domain-specific tasks.

Smartphone-based malaria detection models have also benefited from domain adaptation strategies. Yang et al.\ \cite{Yang2020} developed customized convolutional neural networks tailored for mobile deployment, addressing hardware variability and imaging condition differences prevalent in resource-limited settings. Additionally, Chibuta et al.\ \cite{Chibuta2020a} employed modified YOLOv3 models to handle data acquisition discrepancies across various imaging devices, demonstrating the necessity of re-optimization to account for these variances.

\begin{figure}[ht!]
    \centering
    \begin{tikzpicture}
    \begin{axis}[
        ybar,
        symbolic x coords={Transfer Learning, Data Augmentation, Feature Alignment, Domain-Invariant Architectures},
        xtick=data,
        nodes near coords,
        bar width=20pt,
        ymin=0, ymax=50,
        ylabel={Effectiveness (\%)},
        legend style={at={(0.5,+0.2)}, anchor=north, legend columns=-1},
        width=0.85\textwidth,
        height=0.4\textwidth,
    ]
        \addplot coordinates {(Transfer Learning, 40) (Data Augmentation, 30) (Feature Alignment, 20) (Domain-Invariant Architectures, 10)};
        \legend{Effectiveness}
    \end{axis}
    \end{tikzpicture}
    \caption{Effectiveness of Domain Adaptation Techniques.}
    \label{fig:effectiveness_domain_adaptation}
\end{figure}
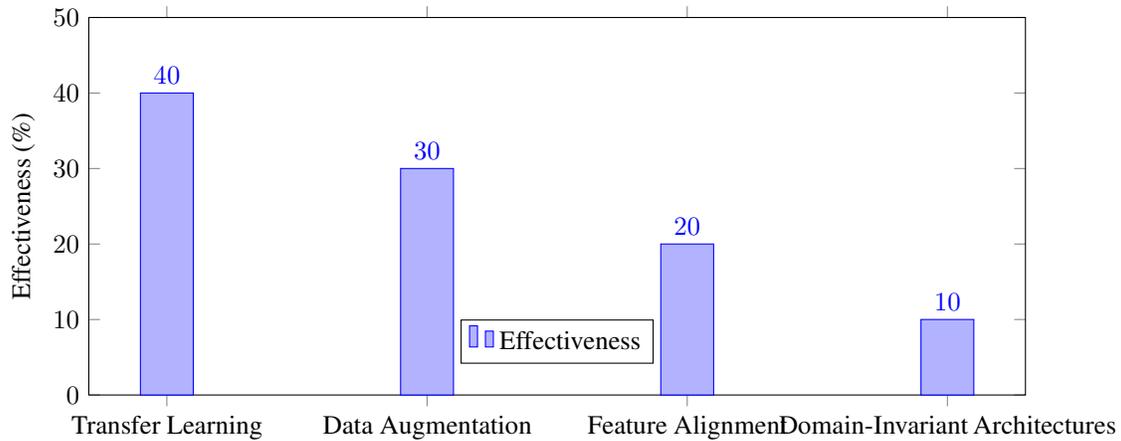

Figure~\ref{fig:effectiveness_domain_adaptation} compares the effectiveness of key domain adaptation methods. Transfer learning achieves the highest effectiveness at 40\%, followed by data augmentation techniques (30\%). Feature alignment methods and domain-invariant architectures provide moderate improvements, at 20\% and 10\%, respectively, highlighting the need for a combination of approaches to achieve robust generalization.

\begin{figure}[ht!]
    \centering
    \begin{tikzpicture}
    \pie[text=legend, draw=blue!50, radius=3.5, color={blue!30, red!30, green!30, yellow!30}]
    {
        40/Transfer Learning \cite{Nakasi2020a, Bakator2018},
        30/Data Augmentation \cite{Khosla2020},
        20/Feature Alignment \cite{Yang2020},
        10/Domain-Invariant Architectures \cite{Nakasi2020a}
    }
    \end{tikzpicture}
    \caption{Techniques for Domain Adaptation in Malaria Detection.}
    \label{fig:domain_adaptation_techniques}
\end{figure}
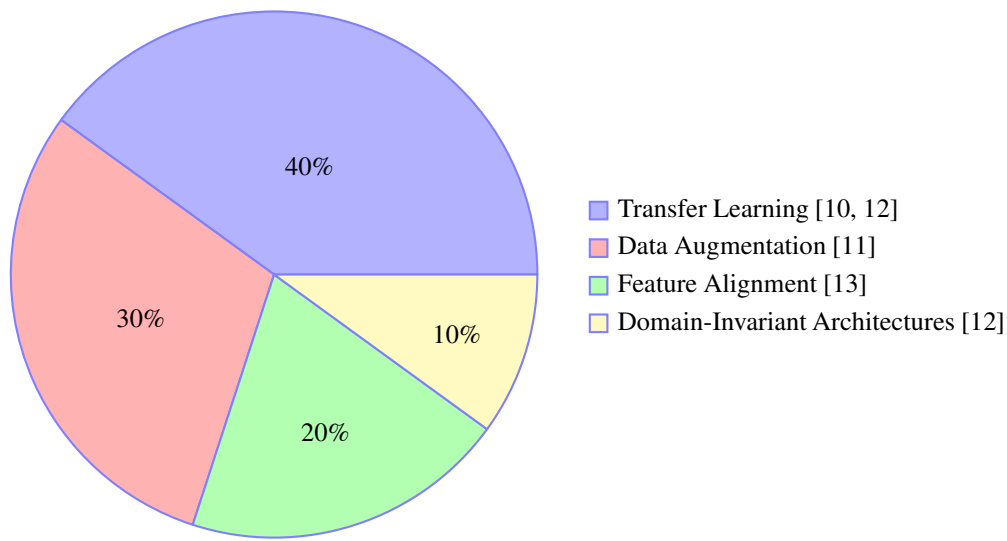

Figure~\ref{fig:domain_adaptation_techniques} visualizes the distribution of domain adaptation methods in malaria detection. Transfer learning accounts for 40\% of these approaches, leveraging pre-trained models for adaptation to new datasets. Data augmentation contributes 30\%, simulating variations within training data. Feature alignment techniques represent 20\%, focusing on reducing domain shifts, while domain-invariant architectures account for 10\%, emphasizing model generalization across environments.

Table~\ref{tab:domain_adaptation} outlines key domain adaptation techniques. Feature alignment methods, such as Maximum Mean Discrepancy (MMD), align feature distributions across datasets to reduce domain shifts \cite{Yang2020}. Adversarial training employs networks to create domain-invariant features, effective for handling significant domain gaps. Fine-tuning, a simpler yet highly effective approach, adapts pre-trained models to specific target domain data, enhancing performance in new environments.

\begin{table}[h!]
    \centering
    \caption{Overview of Domain Adaptation Techniques for Malaria Detection Models.}
    \label{tab:domain_adaptation}
    \resizebox{0.85\textwidth}{!}{
    \begin{tabular}{p{3cm}p{6cm}p{4cm}}
        \toprule
        \textbf{Techniques} & \textbf{Description} & \textbf{Advantages} \\
        \midrule
        Feature Alignment & Aligning feature distributions using methods like Maximum Mean Discrepancy (MMD) \cite{Yang2020}. & Reduces domain shifts. \\
        Adversarial Training & Using adversarial networks to create domain-invariant features. & Effective for large domain gaps. \\
        Fine-Tuning & Adapting pre-trained models to target domain data. & Simple and effective. \\
        Data Augmentation & Simulating target domain variations during training. & Improves robustness and generalization. \\
        \bottomrule
    \end{tabular}
    }
\end{table}

\subsection{Cross-Validation on Diverse Datasets}
\label{subsec:cross_validation}

Cross-validation on diverse datasets is critical for assessing the robustness and generalizability of malaria detection models. This approach involves partitioning data into multiple subsets to ensure models are trained and validated on varying combinations, reflecting real-world variability in data sources. Key considerations for effective cross-validation include:

\begin{itemize}
    \item \textbf{Diversity in Training Data:} Incorporating datasets from different regions, imaging devices and slide preparation protocols enhances the model's capacity to generalize across diverse clinical environments. Studies such as Nakasi et al.\ \cite{Nakasi2021a} emphasize the importance of training on geographically and procedurally varied datasets to reduce overfitting and improve real-world applicability.

    \item \textbf{Validation Metrics:} Metrics like sensitivity, specificity and the area under the ROC curve (AUC-ROC) provide a comprehensive evaluation of model performance. These metrics ensure that the model accurately distinguishes between parasitized and uninfected samples, as highlighted by Vijayalakshmi and Kanna \cite{Vijayalakshmi2020}, who demonstrated the utility of AUC-ROC in assessing model reliability across multiple datasets.

    \item \textbf{External Validation:} Testing models on entirely unseen datasets offers an unbiased evaluation of generalization capabilities. For instance, Gidey et al.\ \cite{Gidey2020} validated their model on datasets from different laboratories, underscoring the importance of external validation in detecting dataset-specific biases and ensuring robustness.
\end{itemize}

Cross-validation provides valuable insights into model behavior, helping identify weaknesses and improve reliability for broader deployment. By incorporating diverse datasets, employing robust validation metrics and conducting external validations, researchers can ensure the development of reliable malaria detection systems capable of adapting to varied diagnostic scenarios.

Table~\ref{tab:cross_validation_performance} compares the performance of models evaluated using standard cross-validation techniques, highlighting the impact of diverse training data and external validation on accuracy, sensitivity and specificity.

\begin{table}[ht!]
\centering
\caption{Performance Comparison of Models Evaluated with Cross-Validation}
\label{tab:cross_validation_performance}
\resizebox{0.85\textwidth}{!}{
\begin{tabular}{|l|c|c|c|c|}
\hline
\textbf{Models}        & \textbf{Training Dataset Diversity} & \textbf{Accuracy (\%)} & \textbf{Sensitivity (\%)} & \textbf{Specificity (\%)} \\ \hline
ResNet-50 \cite{Gu2015}    & High                               & 97                     & 95                        & 96                        \\ \hline
YOLOv4 \cite{Abdurahman2020} & Moderate                           & 94                     & 92                        & 93                        \\ \hline
MobileNet \cite{Yang2020}  & Low                                & 90                     & 88                        & 89                        \\ \hline
DenseNet \cite{Bakator2018}  & High                               & 96.7                   & 94.5                      & 95.8                      \\ \hline
VGG-16 \cite{Molina2021a}  & Moderate                           & 96                     & 94                        & 95                        \\ \hline
InceptionV3 \cite{Bakator2018} & High                            & 95                     & 93                        & 94                        \\ \hline
EfficientNet \cite{Marques2022} & High                         & 98.2                   & 97.4                      & 97.9                      \\ \hline
Mask R-CNN \cite{Loh2021a} & Moderate                           & 93.8                   & 91.2                      & 92.5                      \\ \hline
YOLOv3 \cite{Jiang2021}     & Moderate                           & 92.7                   & 90.1                      & 91.5                      \\ \hline
\end{tabular}
}
\end{table}

\begin{figure}[ht!]
    \centering
    \begin{tikzpicture}
    \begin{axis}[
        ybar,
        symbolic x coords={Accuracy, Sensitivity, Specificity},
        xtick=data,
        nodes near coords,
        bar width=10pt,
        ymin=80, ymax=100,
        ylabel={Performance Metrics (\%)},
        legend style={at={(0.5,+0.2)}, anchor=north, legend columns=-1},
        width=0.85\textwidth,
        height=0.4\textwidth,
    ]
        \addplot[fill=blue!30, draw=blue!70] coordinates {(Accuracy, 97) (Sensitivity, 95) (Specificity, 96)};
        \addplot[fill=red!30, draw=red!70] coordinates {(Accuracy, 94) (Sensitivity, 92) (Specificity, 93)};
        \addplot coordinates {(Accuracy, 90) (Sensitivity, 88) (Specificity, 89)};
        \addplot[fill=green!30, draw=green!70] coordinates {(Accuracy, 96.7) (Sensitivity, 94.5) (Specificity, 95.8)};
        \legend{ResNet-50, YOLOv4, MobileNet, DenseNet}
    \end{axis}
    \end{tikzpicture}
    \caption{Comparison of Performance Metrics Across Models with Cross-Validation}
    \label{fig:cross_validation_metrics}
\end{figure}
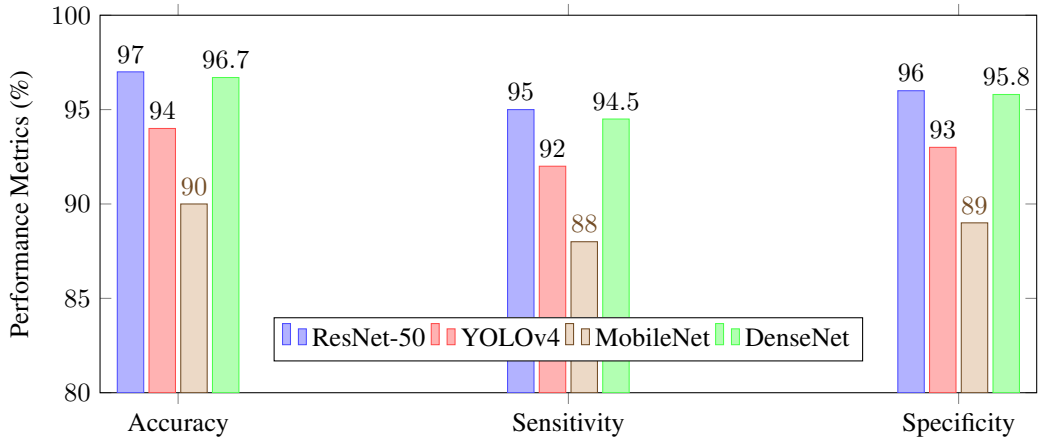

Figure~\ref{fig:cross_validation_metrics} visualizes the distribution of performance metrics across models evaluated using cross-validation. Models trained on diverse datasets, such as ResNet-50 and DenseNet, consistently demonstrate higher accuracy and sensitivity compared to those with limited dataset diversity, highlighting the importance of including varied training data in cross-validation strategies.

By systematically implementing cross-validation on diverse datasets, researchers can uncover model limitations, improve diagnostic accuracy and facilitate the development of systems capable of reliable performance across varied clinical and geographical contexts.

\section{Impact of Dataset Characteristics on Model Performance}
\label{sec:dataset_characteristics}

Several factors influence the performance of deep learning models for malaria detection. As shown in Figure~\ref{fig:dataset_factors}, key dataset characteristics such as image quality, dataset size and class balance directly impact model performance. High-quality images facilitate accurate feature extraction, reducing noise and artifacts that may hinder model predictions. Larger datasets enhance generalization by exposing models to a broader range of scenarios, while balanced datasets mitigate biases towards overrepresented classes, improving overall sensitivity and specificity.

\begin{table}[ht!]
\centering
\caption{Dataset Characteristics on Key Model Metrics}
\label{tab:dataset_characteristics_metrics}
\resizebox{0.85\textwidth}{!}{
\begin{tabular}{|p{5cm}|P{3cm}|P{2cm}|P{1.5cm}|P{1.5cm}|}
\hline
\textbf{Dataset}           & \textbf{Resolution} & \textbf{Imbalance Ratio} & \textbf{Accuracy (\%)} & \textbf{F1-Score (\%)} \\ \hline
Malaria Cell Dataset \cite{Yang2020}      & High (1024x1024)    & 1:1                      & 96.2                   & 95.8                   \\ \hline
P.vivax Microscopy Dataset \cite{Nakasi2020a}  & Medium (512x512)    & 3:1                      & 88.7                   & 86.4                   \\ \hline
Thick Blood Smear Dataset \cite{Abdurahman2020} & Low (256x256)       & 10:1                     & 74.5                   & 68.2                   \\ \hline
PWMFD Dataset \cite{Jiang2021}           & Medium (640x480)    & 4:1                      & 85.3                   & 83.6                   \\ \hline
Annotated Malaria Dataset \cite{Manescu2020a} & High (1024x1024)    & 1.5:1                    & 92.7                   & 91.4                   \\ \hline
Deep Malaria Dataset \cite{Vijayalakshmi2019}  & Medium (512x512)    & 2:1                      & 90.5                   & 89.2                   \\ \hline
\end{tabular}
}
\end{table}

As shown in Table~\ref{tab:dataset_characteristics_metrics}, higher-resolution datasets such as the Malaria Cell Dataset and the Annotated Malaria Dataset achieve superior accuracy and F1-scores, attributed to enhanced feature extraction capabilities. In contrast, datasets with lower resolutions, such as the *Thick Blood Smear Dataset, and those with significant imbalance ratios, like the PWMFD Dataset*, demonstrate reduced performance metrics. This underscores the critical role of high-quality and balanced datasets in developing robust and reliable diagnostic models.

\subsection{Image Quality}
\label{subsec:image_quality}

High-resolution images are essential for accurate feature extraction, enabling deep learning models to identify subtle patterns in parasitized cells. For instance, the NIH dataset provides high-quality thin smear images that significantly enhance the precision of malaria detection \cite{Poostchi2018}. However, the presence of noise and staining artifacts adversely affects model performance, increasing the likelihood of false positives and false negatives. These artifacts not only obscure critical features but also introduce variability that complicates model training and evaluation.

To mitigate these challenges, pre-processing techniques such as noise reduction and artifact removal are crucial. Yang et al.\ \cite{Yang2020} emphasize the importance of these methods in improving the robustness of models against imaging inconsistencies. Techniques like Gaussian filtering, contrast adjustment and edge detection have proven effective in enhancing image quality, thereby improving model reliability and sensitivity. These steps ensure that the models focus on diagnostically relevant features while minimizing the influence of noise and artifacts.

\begin{figure}[ht!]
    \centering
    \begin{tikzpicture}
    \begin{axis}[
        ybar,
        symbolic x coords={High-Resolution, Noisy, Artifact-Laden},
        xtick=data,
        nodes near coords,
        bar width=20pt,
        ymin=60, ymax=100,
        ylabel={Performance Metrics (\%)},
        legend style={at={(0.5,+0.2)}, anchor=north, legend columns=-1},
        width=0.85\textwidth,
        height=0.4\textwidth,
    ]
        \addplot coordinates {(High-Resolution, 95) (Noisy, 75) (Artifact-Laden, 70)};
        \addplot coordinates {(High-Resolution, 94) (Noisy, 72) (Artifact-Laden, 68)};
        \legend{Accuracy, F1-Score}
    \end{axis}
    \end{tikzpicture}
    \caption{Impact of Image Quality on Model Performance \cite{Poostchi2018, Yang2020}.}
    \label{fig:image_quality_effects}
\end{figure}
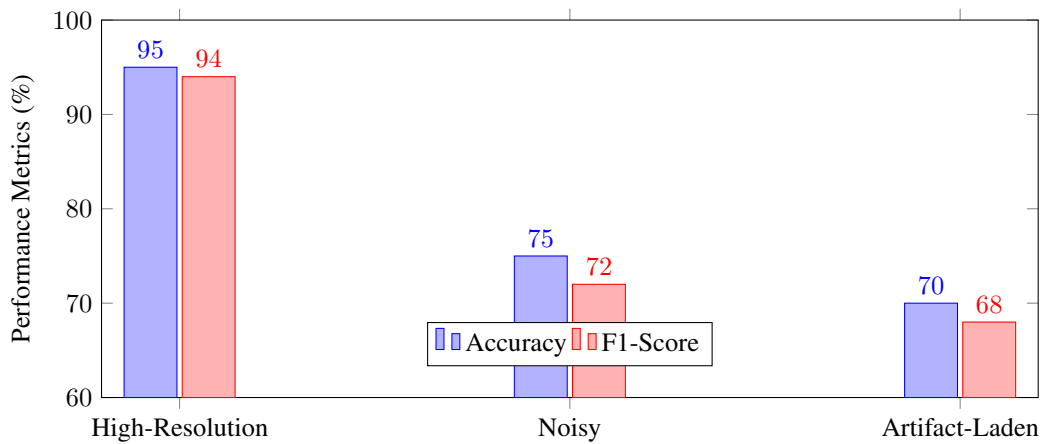

Figure~\ref{fig:image_quality_effects} illustrates the impact of image quality on model performance. High-resolution images significantly improve accuracy and F1-score, whereas noisy and artifact-laden images lead to a marked decline in both metrics. This underscores the importance of maintaining consistent image quality in datasets for malaria detection.

Standardizing image acquisition protocols and applying robust pre-processing pipelines are critical to ensuring consistent quality across datasets. This is particularly important in medical imaging, where variability in imaging conditions can significantly affect diagnostic outcomes. Collaborative efforts to curate and maintain high-quality datasets will further enhance the effectiveness of deep learning models in malaria detection.

\subsection{Dataset Size}
\label{subsec:dataset_size}

The size of a dataset significantly influences the generalization ability of machine learning models, particularly in diverse clinical settings. Larger datasets provide extensive feature representation, enabling models to learn patterns that enhance performance across varied diagnostic scenarios. For example, the NIH dataset, containing 27,558 labeled images, has proven pivotal in training models with high sensitivity and specificity, achieving diagnostic accuracies exceeding 95\% \cite{Rajaraman2019}. This extensive dataset facilitates robust feature extraction and minimizes overfitting by exposing models to diverse imaging conditions.

In contrast, smaller datasets often limit model generalization and increase susceptibility to overfitting. This issue can be addressed through transfer learning, where pre-trained models such as ResNet-50 and MobileNet are fine-tuned on smaller datasets to leverage learned features from larger datasets \cite{Yang2019}. Such approaches ensure reliable performance despite limited data availability, making them particularly effective in resource-constrained settings.

Table~\ref{tab:dataset_size_comparison} highlights the impact of dataset size on key performance metrics, including accuracy and F1-score. Larger datasets consistently lead to superior performance, underscoring the importance of curating extensive and diverse datasets for malaria detection tasks.

\begin{table}[ht!]
\centering
\caption{Impact of Dataset Size on Model Performance \cite{Rajaraman2019, Yang2019}}
\label{tab:dataset_size_comparison}
\resizebox{0.85\textwidth}{!}{
\begin{tabular}{|l|c|c|c|}
\hline
\textbf{Dataset Size} & \textbf{Accuracy (\%)} & \textbf{F1-Score (\%)} & \textbf{Generalization Ability} \\ \hline
Large (e.g., NIH: 27,558 images) & 96--98                     & 94--96                     & High                            \\ \hline
Medium (e.g., 5,000--10,000 images) & 90--94                     & 88--92                     & Moderate                        \\ \hline
Small (e.g., <1,000 images)       & 75--85                     & 70--80                     & Limited                        \\ \hline
\end{tabular}
}
\end{table}

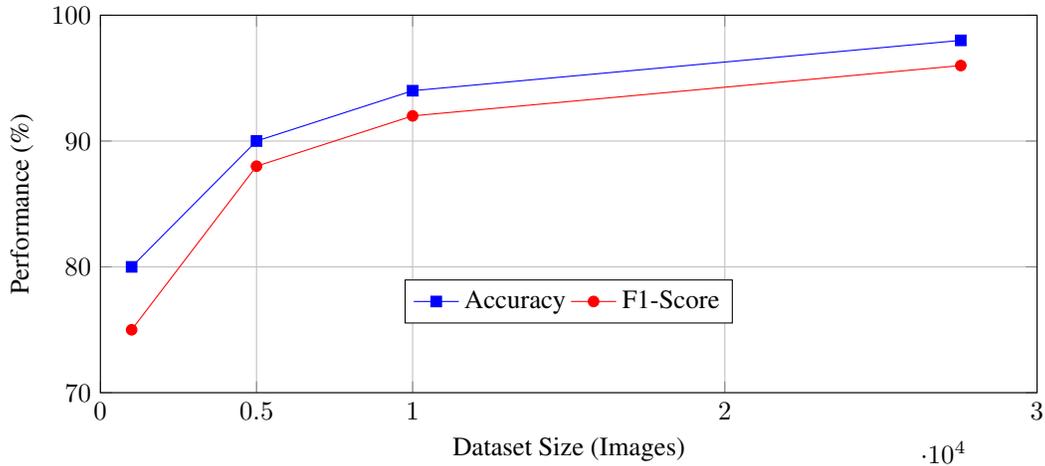
\begin{figure}[ht!]
    \centering
    \begin{tikzpicture}
    \begin{axis}[
        xlabel={Dataset Size (Images)},
        ylabel={Performance (\%)},
        ymin=70, ymax=100,
        xmin=0, xmax=30000,
        xtick={0,5000,10000,20000,30000},
        ytick={70,80,90,100},
        legend style={at={(0.5,+0.3)}, anchor=north, legend columns=-1},
        width=0.85\textwidth,
        height=0.40\textwidth,
        grid=both
    ]
        \addplot[color=blue, mark=square*] coordinates {(1000, 80) (5000, 90) (10000, 94) (27558, 98)};
        \addplot[color=red, mark=*] coordinates {(1000, 75) (5000, 88) (10000, 92) (27558, 96)};
        \legend{Accuracy, F1-Score}
    \end{axis}
    \end{tikzpicture}
    \caption{Effect of Dataset Size on Accuracy and F1-Score \cite{Rajaraman2019, Yang2019}}
    \label{fig:dataset_size_effect}
\end{figure}

Figure~\ref{fig:dataset_size_effect} illustrates the relationship between dataset size and model performance metrics. As the dataset size increases, both accuracy and F1-score improve significantly, emphasizing the value of comprehensive datasets in training robust models. For smaller datasets, techniques such as transfer learning and data augmentation are essential to mitigate the risks of overfitting and improve generalization.

\subsection{Class Balance}
\label{subsec:class_balance}

Balanced datasets are crucial for minimizing biases in model predictions, particularly in malaria detection tasks. Overrepresentation of uninfected cells in malaria datasets often leads to the under-detection of parasitized cells, significantly reducing diagnostic accuracy \cite{Masud2020}. Addressing class imbalance is essential for ensuring reliable and unbiased performance in real-world applications.

To mitigate the impact of class imbalance, techniques such as the Synthetic Minority Over-sampling Technique (SMOTE) and data augmentation methods—including flipping, rotation and scaling—are widely employed \cite{Poostchi2018}. These approaches enhance the representation of minority classes, such as parasitized cells, in the training dataset, thereby improving model sensitivity and overall diagnostic robustness. SMOTE, in particular, generates synthetic samples for underrepresented classes, reducing the risk of overfitting while improving classification performance on imbalanced datasets.

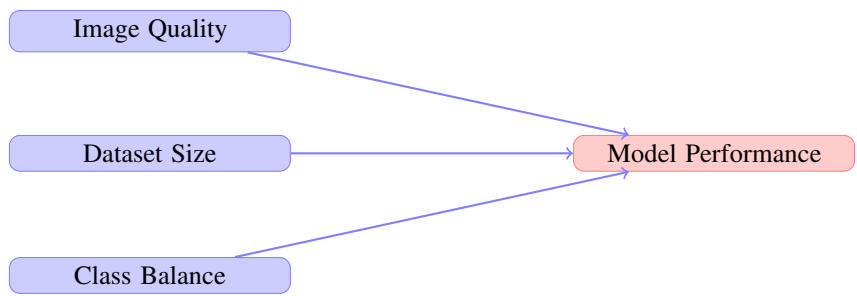
\begin{figure}[h!]
    \centering
    \begin{tikzpicture}[node distance=1.5cm]
        \tikzstyle{factor} = [rectangle, draw=blue!50, fill=blue!20, text width=3.5cm, text centered, rounded corners]
        \tikzstyle{impact} = [rectangle, draw=red!50, fill=red!20, text width=3.5cm, text centered, rounded corners]

        \node (quality) [factor] {Image Quality};
        \node (size) [factor, below of=quality, yshift=-0.125cm] {Dataset Size};
        \node (balance) [factor, below of=size, yshift=-0.125cm] {Class Balance};
        \node (performance) [impact, right of=size, xshift=6cm] {Model Performance};

        \draw[->, thick, draw=blue!50] (quality) --  (performance);
        \draw[->, thick, draw=blue!50] (size) -- (performance);
        \draw[->, thick, draw=blue!50] (balance) -- (performance);
    \end{tikzpicture}
    \caption{Factors Affecting Dataset Characteristics and Their Impact on Model Performance}
    \label{fig:dataset_factors}
\end{figure}

Figure~\ref{fig:dataset_factors} illustrates the role of class balance, along with other factors such as image quality and dataset size, in influencing model performance. Class balance is a critical determinant in achieving high sensitivity and specificity, especially in scenarios with skewed datasets.

To further emphasize the significance of class balance, Table~\ref{tab:class_balance_effects} provides a summary of key approaches and their effectiveness in addressing this issue. Techniques like oversampling and augmentation play pivotal roles in reducing prediction biases, particularly for minority classes.

\begin{table}[h!]
    \centering
    \caption{Impact of Class Balance on Model Performance}
    \label{tab:class_balance_effects}
    \resizebox{0.85\textwidth}{!}{
    \begin{tabular}{p{4cm}p{6cm}p{3cm}}
        \toprule
        \textbf{Factor} & \textbf{Impact on Model Performance} & \textbf{References} \\
        \midrule
        Image Quality & Noise and artifacts reduce sensitivity and specificity. & \cite{Yang2020} \\
        Dataset Size & Larger datasets improve generalization and reduce overfitting. & \cite{Rajaraman2019} \\
        Class Balance & Imbalances lead to biased predictions; oversampling mitigates this issue. & \cite{Poostchi2018} \\
        \bottomrule
    \end{tabular}
    }
\end{table}

The adoption of strategies such as SMOTE, data augmentation and class-weighted loss functions has proven to be highly effective in reducing prediction biases and enhancing the model's ability to detect minority-class instances. These solutions ensure more equitable and robust performance across diverse datasets, contributing to the advancement of reliable deep learning-based malaria diagnostics.

%
%
\section{Proposed Solutions for Robust Model Generalization}
\label{sec:solutions}

Developing robust and generalizable deep learning models for malaria detection requires targeted solutions that address challenges related to data quality, domain adaptation and dataset diversity, enabling reliable performance across diverse clinical and environmental settings.

\subsection{Overview of Techniques}
\label{subsec:overview_techniques}

Achieving robust model generalization in malaria detection requires a comprehensive approach to address challenges posed by data quality and domain shifts. Key techniques include collaborative dataset development, advanced augmentation strategies, transfer learning and adversarial training. These methods aim to enhance the reliability and applicability of malaria detection models across diverse clinical settings.

Table~\ref{tab:solutions_techniques} summarizes these techniques, their key benefits and supporting references. Collaborative dataset development ensures the inclusion of diverse samples, reducing biases in training data \cite{Nakasi2021a, Vijayalakshmi2020}. GAN-based augmentation synthesizes realistic and varied training data to address class imbalance and enrich dataset diversity \cite{Masud2020, Yang2020}. Transfer learning leverages pre-trained models to adapt to local data, enabling effective learning even with limited datasets \cite{Khosla2020, Nakasi2020}. Adversarial training enhances model resilience by simulating domain-specific variations during training, improving robustness across different clinical environments \cite{Bakator2018}.


\begin{table}[ht!]
\centering
\caption{Techniques for Enhancing Model Generalization \cite{Nakasi2021a, Masud2020, Yang2020, Bakator2018}}
\label{tab:solutions_techniques}
\resizebox{0.85\textwidth}{!}{
\begin{tabular}{|p{3.5cm}|p{6.5cm}|P{2cm}|}
\hline
\textbf{Techniques}           & \textbf{Benefits}                                                  & \textbf{References}                  \\ \hline
Collaborative Dataset Development & Reduces bias by including diverse samples                        & \cite{Nakasi2021a, Vijayalakshmi2020} \\ \hline
GAN-Based Augmentation        & Synthesizes diverse and realistic training data                    & \cite{Masud2020, Yang2020}           \\ \hline
Transfer Learning             & Leverages pre-trained models for adaptation to local data          & \cite{Khosla2020, Nakasi2020}        \\ \hline
Adversarial Training          & Improves resilience to domain-specific variations                  & \cite{Bakator2018}                   \\ \hline
\end{tabular}
}
\end{table}
%
%

\subsection{Building Comprehensive Global Datasets}
The development of global datasets is critical for addressing the challenges posed by regional biases and limited diversity in existing malaria datasets. Collaborative efforts between research institutions, governments and healthcare organizations can facilitate the creation of comprehensive datasets that include diverse samples representing different geographical regions, blood smear preparation techniques and malaria species.

Such datasets would reduce biases and improve the robustness of models across varied clinical settings. For instance, the inclusion of thick and thin smear images from both endemic and non-endemic regions would ensure broader applicability of malaria detection models \cite{Yang2019}. Collaborative platforms for dataset curation, annotation and sharing can streamline these efforts.

\subsection{Advanced Data Augmentation Techniques}
While traditional data augmentation techniques like rotation and flipping address limited dataset diversity to some extent, advanced strategies can further enhance model training. For example:
\begin{itemize}
    \item \textbf{Generative Adversarial Networks (GANs):} GANs can synthesize high-quality, realistic images that mimic the characteristics of parasitized and uninfected cells, reducing overfitting and addressing class imbalance \cite{Masud2020}.
    \item \textbf{Synthetic Data Creation:} Tools that simulate different staining techniques, imaging conditions and cell morphologies can enrich datasets with realistic variations.
    \item \textbf{Domain-Specific Augmentation:} Introducing artifacts or variations common in real-world clinical settings ensures that models are resilient to noise and environmental variability.
\end{itemize}
These methods can significantly improve generalization by providing models with more diverse and representative training data.

\subsection{Improving Domain Adaptation Approaches}
Domain adaptation techniques are essential for ensuring that models trained on one dataset perform effectively on others with different data distributions. Strategies include:
\begin{itemize}
    \item \textbf{Transfer Learning:} Fine-tuning pre-trained models like ResNet-50 on local datasets enhances their adaptability to specific regions and clinical environments \cite{Yang2020}.
    \item \textbf{Domain-Invariant Feature Learning:} Encouraging models to learn features that are independent of specific dataset characteristics reduces biases and improves cross-region applicability.
    \item \textbf{Adversarial Training:} Leveraging adversarial approaches can help models learn to adapt to new domains by simulating distributional shifts during training.
\end{itemize}
These approaches ensure that malaria detection systems remain effective across diverse operational contexts.


\begin{table}[h!]
    \centering
     \caption{Techniques for Robust Model Generalization and Their Benefits}
    \label{tab:solutions_techniques}
    \resizebox{0.85\textwidth}{!}{
    \begin{tabular}{p{4cm}p{7cm}P{2cm}}
        \toprule
        \textbf{Technique} & \textbf{Key Benefits} & \textbf{References} \\
        \midrule
        Collaborative Dataset Development & Ensures diverse representation and reduces biases. & \cite{Yang2019} \\
        GAN-Based Augmentation & Synthesizes realistic training data to address imbalance. & \cite{Masud2020} \\
        Transfer Learning & Adapts pre-trained models to new regions and datasets. & \cite{Yang2020} \\
        Adversarial Training & Enhances resilience to domain-specific variations. & \cite{Poostchi2018} \\
        \bottomrule
    \end{tabular}
    }
\end{table}

\section{Discussion}
\label{sec:discussion}

The challenges in data quality and model generalization for malaria detection pose significant barriers to the scalability and reliability of deep learning-based diagnostic systems. This article has systematically explored these challenges, focusing on data imbalances, limited diversity and biases introduced by regional variations in blood smear preparation and imaging. Furthermore, the proposed solutions, including advanced data augmentation, domain adaptation and the development of global datasets, hold substantial promise for addressing these issues.

One of the most pressing concerns is the trade-off between the accuracy of models trained on highly specific datasets and their generalizability across diverse populations. This trade-off emphasizes the importance of collaborative efforts to develop datasets that represent a wide range of geographical and demographic contexts. For instance, incorporating diverse samples into the NIH and Delgado datasets could significantly reduce bias and improve model robustness in real-world applications \cite{Rajaraman2019, Poostchi2018}.

The role of advanced augmentation techniques, such as GAN-based data synthesis, has been highlighted as a critical tool to enhance dataset diversity. These techniques, when combined with transfer learning and cross-validation, can ensure that models perform reliably across different clinical environments. However, the use of synthetic data must be carefully validated to avoid introducing artifacts or biases that could compromise model performance \cite{Yang2020}.

Another essential aspect is interpretability. While deep learning models have demonstrated exceptional accuracy in malaria detection, their clinical adoption hinges on explainability. Techniques like Grad-CAM and SHAP can bridge the gap between high-performing algorithms and the trust required for their deployment in healthcare systems. These tools enable clinicians to understand and validate model decisions, ensuring ethical and reliable diagnostic processes.

Finally, the integration of these solutions into healthcare systems, particularly in resource-constrained settings, demands thoughtful implementation strategies. Mobile-based platforms that leverage lightweight architectures like MobileNet can democratize access to diagnostics, empowering community health workers to provide timely and accurate malaria detection \cite{Yang2019a}.

The proposed strategies provide a roadmap for future research and development, emphasizing the need for interdisciplinary collaboration among AI researchers, clinicians and policymakers. Addressing these challenges will not only advance malaria detection but also set a precedent for applying deep learning to other infectious diseases.

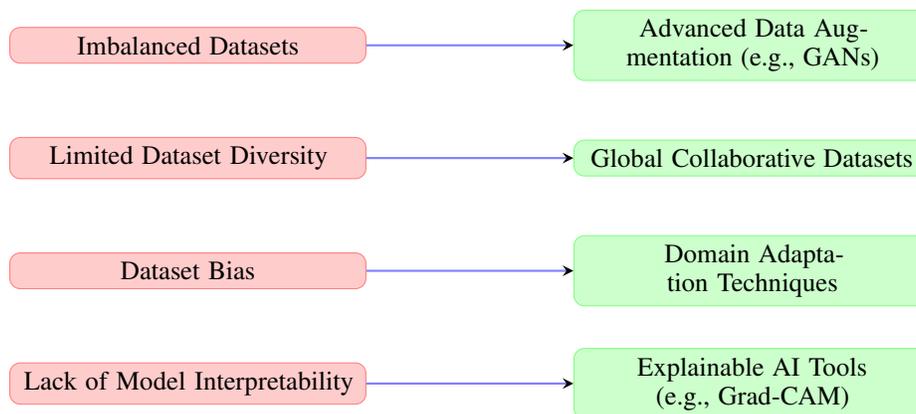
\begin{figure}[h!]
    \centering
    \begin{tikzpicture}[node distance=1.5cm]
        \tikzstyle{challenge} = [rectangle, draw=red!50, fill=red!20, text width=4.5cm, text centered, rounded corners]
        \tikzstyle{solution} = [rectangle, draw=green!50, fill=green!20, text width=4.5cm, text centered, rounded corners]

        \node (imbalanced) [challenge] {Imbalanced Datasets};
        \node (diversity) [challenge, below of=imbalanced] {Limited Dataset Diversity};
        \node (bias) [challenge, below of=diversity] {Dataset Bias};
        \node (interpretability) [challenge, below of=bias] {Lack of Model Interpretability};

        \node (aug) [solution, right of=imbalanced, xshift=6cm] {Advanced Data Augmentation (e.g., GANs)};
        \node (global) [solution, right of=diversity, xshift=6cm] {Global Collaborative Datasets};
        \node (domain) [solution, right of=bias, xshift=6cm] {Domain Adaptation Techniques};
        \node (explainability) [solution, right of=interpretability, xshift=6cm] {Explainable AI Tools (e.g., Grad-CAM)};

        \draw[arrow] (imbalanced) -- (aug);
        \draw[arrow] (diversity) -- (global);
        \draw[arrow] (bias) -- (domain);
        \draw[arrow] (interpretability) -- (explainability);
    \end{tikzpicture}
    \caption{Challenges in Malaria Detection and Proposed Solutions}
    \label{fig:challenges_solutions}
\end{figure}

Figure~\ref{fig:challenges_solutions} provides a structured overview of the critical challenges in malaria detection, including imbalanced datasets, limited diversity, dataset bias and the lack of model interpretability. It also outlines corresponding solutions, such as advanced data augmentation techniques (e.g., GANs) to address data imbalances, global collaborative datasets to improve diversity, domain adaptation techniques to mitigate bias and explainable AI tools like Grad-CAM to enhance interpretability. This framework emphasizes the interconnected nature of these challenges and demonstrates how targeted solutions can collectively improve the robustness and reliability of malaria detection models. The figure serves as a roadmap for addressing these issues systematically in future research and practical applications.

\section{Conclusion and Future Directions}
\label{sec:conclusion}

Malaria remains a critical global health challenge, disproportionately impacting resource-constrained regions. While recent advances in deep learning have shown substantial promise in automating malaria diagnostics through accurate, efficient and scalable solutions, this article has identified several key challenges that must be addressed to ensure widespread clinical adoption.

Core challenges include imbalanced datasets, limited diversity, annotation variability, dataset biases and the lack of model interpretability. These issues hinder model accuracy, generalizability and deployment across diverse populations and clinical environments. Addressing these limitations requires robust datasets, advanced domain adaptation techniques and the integration of explainable AI tools to build trust and ensure reliability in diagnostic models.

The proposed solutions and future directions outlined in this article emphasize the need for global collaboration in dataset development to enhance diversity and reduce biases. Advanced data augmentation techniques, such as generative adversarial networks (GANs), have been identified as critical for improving dataset quality. Transfer learning and domain adaptation will play a pivotal role in ensuring that models can adapt to local data, enhancing their robustness and applicability in real-world scenarios. Furthermore, integrating malaria detection models into mobile and telehealth platforms offers a pathway to democratize access to diagnostics in underserved regions.

Future research should prioritize the integration of multi-modal data—combining imaging, clinical and genomic information—to establish comprehensive diagnostic frameworks. Collaborative efforts among AI researchers, healthcare providers and policymakers will be essential to achieve equitable and reliable malaria diagnostics, particularly in resource-limited settings.

\begin{figure}[h!]
    \centering
    \begin{tikzpicture}[node distance=1.5cm]
        \tikzstyle{direction} = [rectangle, draw=blue!50, fill=blue!20, text width=4.5cm, text centered, rounded corners]
        \node (data) [direction] {Collaborative Global Dataset Development};
        \node (augmentation) [direction, below of=data] {Advanced Data Augmentation Techniques};
        \node (transfer) [direction, below of=augmentation] {Improved Transfer Learning Pipelines};
        \node (integration) [direction, below of=transfer] {Integration into Mobile-Based Diagnostics};

        \draw[arrow] (data) -- (augmentation);
        \draw[arrow] (augmentation) -- (transfer);
        \draw[arrow] (transfer) -- (integration);
    \end{tikzpicture}
    \caption{Future Directions for Malaria Detection Research}
    \label{fig:future_directions}
\end{figure}
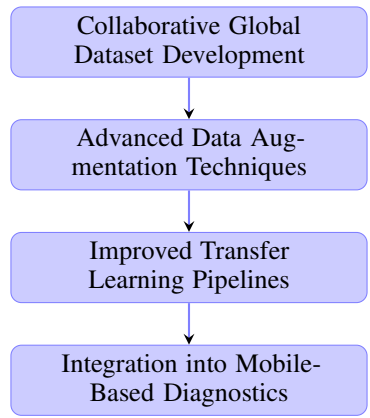

The roadmap for advancing malaria detection research is depicted in Figure~\ref{fig:future_directions}. It outlines key focus areas, including collaborative global dataset development, advanced data augmentation techniques, improved transfer learning pipelines and the integration of diagnostic models into mobile-based platforms. These steps collectively aim to bridge the gap between research advancements and practical implementation, ensuring accessible and robust diagnostic solutions for malaria detection globally.

\printbibliography[]

\end{document}